\documentclass[letterpaper]{article} 
\usepackage{aaai25}  
\usepackage{times}  
\usepackage{helvet}  
\usepackage{courier}  
\usepackage[hyphens]{url}  
\usepackage{graphicx} 
\usepackage{natbib}  
\usepackage{caption} 
\usepackage{algorithm}
\usepackage{algorithmic}

\usepackage{newfloat}
\usepackage{listings}
\DeclareCaptionStyle{ruled}{labelfont=normalfont,labelsep=colon,strut=off} 
\floatstyle{ruled}
\newfloat{listing}{tb}{lst}{}
\floatname{listing}{Listing}
\usepackage{amsmath}
\usepackage{amsfonts}
\usepackage{xcolor}
\usepackage{dsfont}

\nocopyright 

\title{Learning Production Functions for Supply Chains with Graph Neural Networks\thanks{This is the extended version of a paper accepted to AAAI'25.}}
\author{
    Serina Chang\textsuperscript{\rm 1},
    Zhiyin Lin\textsuperscript{\rm 1},
    Benjamin Yan\textsuperscript{\rm 1},
    Swapnil Bembde\textsuperscript{\rm 2},
    Qi Xiu\textsuperscript{\rm 2},
    Chi Heem Wong\textsuperscript{\rm 1,2},\\
    Yu Qin\textsuperscript{\rm 2,3},
    Frank Kloster\textsuperscript{\rm 2},
    Xi Luo\textsuperscript{\rm 2},
    Raj Palleti\textsuperscript{\rm 1,2},
    Jure Leskovec\textsuperscript{\rm 1}
}
\affiliations{
    \textsuperscript{\rm 1}Stanford University, Department of Computer Science\\
   \textsuperscript{\rm 2}Hitachi America, Ltd.\\
   \textsuperscript{\rm 3}Tulane University\\
}

\begin{document}

\maketitle

\begin{abstract}
The global economy relies on the flow of goods over supply chain networks, with nodes as firms and edges as transactions between firms.
  While we may observe these external transactions, they are governed by unseen \textit{production functions}, which determine how firms internally transform the input products they receive into output products that they sell. 
  In this setting, it can be extremely valuable to infer these production functions, to improve supply chain visibility and to forecast future transactions more accurately.
  However, existing graph neural networks (GNNs) cannot capture these hidden relationships between nodes' inputs and outputs.
  Here, we introduce a new class of models for this setting by combining temporal GNNs with a novel inventory module, which learns production functions via attention weights and a special loss function.
  We evaluate our models extensively on real supply chains data and data generated from our new open-source simulator, \texttt{SupplySim}. 
  Our models successfully infer production functions, outperforming the strongest baseline by 6\%--50\% (across datasets), and forecast future transactions, outperforming the strongest baseline by 11\%--62\%.
\end{abstract}

%

\newcommand{\serina}[1]{{{\textcolor{blue}{[Serina: #1]}}}}
\newcommand{\frank}[1]{{{\textcolor{green}{[Frank: #1]}}}}
\newcommand{\swapnil}[1]{{{\textcolor{orange}{[Swapnil: #1]}}}}

\newcommand{\inv}[1]{\mathbf{x}_{#1}^{(t)}}
\newcommand{\emb}[1]{\mathbf{z}_{#1}^{(t)}}
\newcommand{\mem}[1]{\mathbf{m}_{#1}^{(t)}}
\newcommand{\github}{\url{https://github.com/snap-stanford/supply-chains/tree/tgb}}

\newcommand{\yes}{{{\textcolor{blue}{[Yes]}}}}
\newcommand{\partialcheck}{{{\textcolor{purple}{[Partial]}}}}
\newcommand{\no}{{{\textcolor{orange}{[No]}}}}
\newcommand{\NA}{{{\textcolor{gray}{[NA]}}}}
\section{Introduction}
Supply chains form the backbone of the global economy and disruptions can have enormous consequences, costing trillions of dollars \citep{baumgartner2020shocks} and risking national security \citep{whitehouse2021}.
Thus, modeling supply chains and how they evolve, especially under shocks, is essential.
Prior models of supply chains have been mainly mechanistic and struggle to fit even highly aggregated measures, such as country-level production \citep{inoue2019japan}.
As a result, recent literature has called for integration of machine learning (ML) into supply chain modeling \citep{baryannis2019ml,brintrup2020data,younis2022ai}.

Graph neural networks (GNNs) serve as a particularly promising ML methodology, as supply chains are naturally represented as graphs, with nodes as firms and edges as transactions between firms.
A few works have explored static GNNs for supply chains \citep{aziz2021grl,kosasih2021gnn,wasi2024supplygraph}, but these static models miss crucial dynamic aspects of supply chains, such as those driving the propagation of shocks.
Thus, there is a need to develop temporal GNNs that can capture the unique dynamics and mechanisms of supply chain graphs.

One key feature of these graphs is that they are governed by underlying \textit{production functions}: firms receive input products (e.g., wheels) from other firms, internally transform those inputs into outputs (e.g., cars) via production functions, then sell those output products to other firms or consumers \citep{carvalho2019primer}.
These production functions dictate which firms are connected to each other, as well as the timing of transactions across the network. 
For example, if a firm experiences a shortage in an input, its production of outputs that rely on that input will be disrupted, but it can continue to produce other outputs for some time.
However, existing temporal GNNs \citep{huang2023tgb} are not designed to learn these hidden production functions or to incorporate them into link prediction.

\textbf{The present work.}
Here, we are the first to develop temporal GNNs for supply chains, demonstrating their ability to learn rich, dynamic representations of firms and products.
Furthermore, by introducing supply chains as a new application, we identify a challenging setting unexplored by prior GNNs: temporal graphs governed by production functions, which we term \textit{temporal production graphs} (TPGs).
Beyond supply chains, TPGs also appear in biological domains, such as enzymes producing metabolites within metabolic pathways, or in organizations, where teams rely on inputs from other teams to produce their outputs (e.g., engineers interfacing with product designers). 
TPGs introduce new challenges compared to other temporal graphs, such as the need to infer production functions.
Furthermore, even standard tasks, like link prediction, may be complicated under TPGs.
For example, while standard temporal GNNs might capture disruptions generally propagating across connected firms, they will miss the specific connections between each firm's inputs and outputs and, under a shortage of inputs, not be able to precisely predict which outputs will be affected.

Since existing GNNs cannot handle TPGs, we introduce a \textit{new class of GNNs} designed for TPGs, focusing on two objectives:  (1) learning the graph's production functions, (2) predicting its future edges.
Prior temporal GNNs have focused primarily on the second objective, but are not designed for the first.
Our models support both objectives, by combining temporal GNNs with a novel inventory module, which learns production functions by explicitly representing each firm's inventory and updating it based on attention weights that map external supply to internal consumption. 
Our module can be combined with \textit{any} GNN; to demonstrate this, here we combine it with two popular models, Temporal Graph Network \citep{rossi2020tgn} and GraphMixer \citep{cong2023graphmixer}, which we also extend in important ways.

We demonstrate the utility of our models on real and synthetic supply chains data.
Through an industry collaboration, we have rare access to real \textit{transaction-level} supply chains data, which allows us to evaluate our GNNs' abilities to predict individual, future transactions.
However, since we cannot release the proprietary data, we also build a new open-source simulator, \texttt{SupplySim}, which generates realistic supply chain data that matches the real data on key characteristics. 
Our simulator enables us to share data, test models under controlled and varied settings (e.g., supply shocks, missing data), and encourage future ML research on supply chains and TPGs.
In summary, our contributions are:
\begin{enumerate}
    \item \textbf{Problem}: a new graph ML problem setting, \textit{temporal production graphs} (TPGs), where each node's edges are related via unobserved production functions,
    \item \textbf{Models}: a new class of models for TPGs, which combine temporal GNNs with a novel inventory module to jointly learn production functions and forecast future edges,
    \item \textbf{Data}: an open-source simulator, \texttt{SupplySim}, which generates realistic supply chain data and enables model testing under a wide variety of settings,
    \item \textbf{Results}: experiments on real and synthetic data, showing that our models effectively learn production functions (outperforming baselines by 6-50\%) and forecast future edges (outperforming baselines by 11-62\%).
\end{enumerate}
Our work both contributes to the supply chains domain, as the first to develop temporal GNNs for supply chains, and advances graph ML, by introducing a new type of real-world graph and developing new methods to handle them.\footnote{Our code to run all experiments, simulator \texttt{SupplySim}, and synthetic datasets are available at \github.}

\paragraph{Social impact.}
Our work is a collaboration with Hitachi America, Ltd., a multinational conglomerate that produces a wide range of products, each involving a complex global supply chain.
Our project was initiated by the company's need for graph-based ML solutions for supply chains.
By working together, we identified two ML objectives grounded in real business needs.
Our first objective of learning production functions aims to improve supply chain visibility, so that firms can better understand the entire production process through which their products are produced (instead of only their immediate suppliers), identify bottlenecks, and find more efficient solutions \citep{aigner1968prod,coelli2005efficiency}. 
Our second objective of forecasting future transactions supports critical industry tasks, such as demand forecasting \citep{seyedan2020demand}, early detection of risks \citep{sheffi2015detection}, and inventory optimization \citep{vandeput2020inventory}.
Additionally, our collaboration provides access to rare, transactions-level data, which enables us to rigorously evaluate our ML models and build a realistic simulator so that, despite the data sharing constraints of this domain, we can still encourage future ML research on supply chains.
\section{Related Work}
Many real-world systems can be represented as temporal graphs, such as transportation systems \citep{jiang2022traffic}, human mobility \citep{chang2023spillovers}, and biological networks \citep{prill2005bio}.
While most GNNs are designed for static graphs, there has been growing interest recently in developing GNNs for temporal graphs \citep{huang2023tgb,skarding2021survey,longa2023survey}.
However, to the best of our knowledge, GNNs have not yet been designed for temporal production graphs (TPGs), as described in this work.
Furthermore, only a handful of works have explored GNNs for supply chains, all in static settings, such as to predict hidden links between firms \citep{aziz2021grl,kosasih2021gnn}, classify a firm's industry \citep{wu2023classification}, recommend suppliers \citep{tu2024rec}, and predict product relations (e.g., same product group) \citep{wasi2024supplygraph}. 


Within the supply chains domain, our work builds on prior literature that represents supply chains as networks \citep{fujiwara2010structure} and models the propagation of shocks over these networks \citep{acemoglu2012fluctuations,zhao2019adaptive,li2020network,carvalho2021qje}.
Our work is unique in two key ways: first, much of the prior work relies on theoretical models and synthetic networks, and, even among empirical studies, the data used is typically industry-level or, at best, firm-level with static relations between firms \citep{carvalho2019primer}.
In contrast, we have access to \textit{transaction}-level data, which reveals essential time-varying information. 
Second, integration of ML into supply chains has been limited, with calls for further exploration \citep{baryannis2019ml,brintrup2020data,younis2022ai}.
Most prior models for modeling supply chains are mechanistic \citep{hallegatte2008ario,guan2020covid,inoue2020covid,li2021ripple} and limited in their ability to even fit aggregate counts \citep{inoue2019japan}.
In contrast, by combining our inventory module with GNNs, we can accurately forecast individual transactions, while maintaining the interpretability of the inventory module.

This theme of combining deep learning with domain-specific principles also appears in physics simulation \citep{wang2021physics} and epidemiological forecasting \citep{liu2024epi}.
Our work also has connections to temporal causal discovery \citep{nauta2019attention,löwe2022acd,assaad2022survey} and inferring networks from node marginals \citep{kumar2015wsdm,maystre2017choicerank,chang2024ipf}. 
\section{Learning on Temporal Production Graphs}
\subsection{Problem definition}

\begin{figure*}
    \centering
    \includegraphics[width=0.9\linewidth]{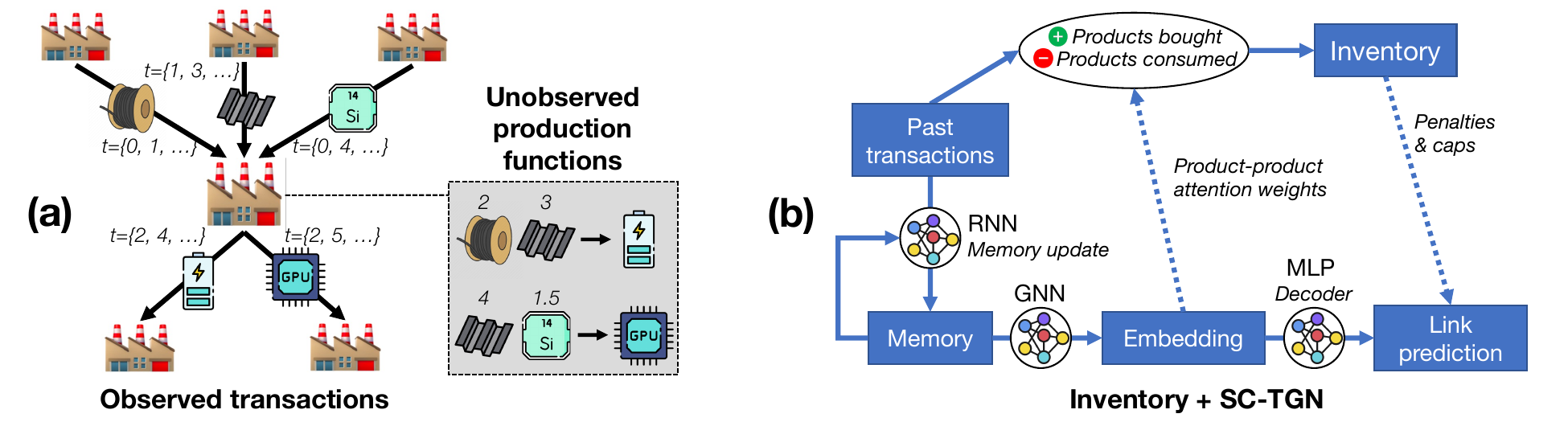}
    \caption{\textbf{(a)} Illustration of our problem setting: we observe time-varying transactions between firms and do not observe production functions within firms. Our goals are to learn the production functions and predict future transactions. \textbf{(b)} Example of our model architecture, combining our inventory module with our extended version of TGN, SC-TGN.} 
    \label{fig:fig1}
\end{figure*}

We define a new graph ML setting called \textit{temporal production graphs} (TPGs), illustrated in Figure \ref{fig:fig1}a. 
TPGs are directed graphs with time-varying edges and potentially a time-varying set of nodes.
What differentiates TPGs from other temporal graphs is that, in a TPG, each node's in-edges represent inputs to some internal production function, which are transformed into outputs, represented by the node's out-edges.
In this work, we focus on supply chain networks, as a canonical example of TPGs.
We are given a set of transactions $\mathcal{T}$ between firms where, for each transaction, we know its timestamp, supplier firm, buyer firm, product sold, and amount sold.
We represent this data as a heterogeneous temporal graph, $G_{\textrm{txns}} = \{\mathcal{N}, \mathcal{E}\}$, where the nodes $\mathcal{N}$ consist of $n$ firm nodes and $m$ product nodes and the edges are $\mathcal{E} := \{e(s, b, p, t)\}$, where $e(s, b, p, t)$ is a \textit{hyperedge} between supplier firm $s$, buyer firm $b$, and product $p$, representing a transaction between them at time $t$.

In this setting, production functions define how firms internally transform the products that they buy into products that they supply. 
Specifically, the function $\mathcal{F}_p: \mathbb{R}_+ \rightarrow \mathbb{R}^m_+$ for product $p$ defines how much of each product is necessary to make $k$ amount of $p$ (e.g., a car requires four wheels).
The set of production functions define a \textit{production graph}, $G_{\textrm{prod}}$, which is a directed acyclic graph where there is an edge from products $p_1$ to $p_2$ if $p_1$ is required to make $p_2$.\footnote{For simplicity, we assume in this work that there is one way to make each product, but future work may consider extensions, such as firm-specific product graphs or substitute products.}
Given the set of transactions $\mathcal{T}$, and the resulting temporal graph $G_{\textrm{txns}}$, our goals are two-fold: (1) to infer $G_{\textrm{prod}}$, which is entirely unobserved, (2) to predict future transactions, i.e., future hyperedges in $G_{\textrm{txns}}$.


\subsection{Model architecture} 
To learn from TPGs, we introduce a new class of models that combine temporal GNNs with a novel inventory module to jointly learn production functions and predict future edges.
First, we describe our inventory module, which can either operate as a stand-alone model or be attached to any GNN.
Second, we describe extended versions of two popular temporal GNNs, Temporal Graph Network \citep{rossi2020tgn} and GraphMixer \citep{cong2023graphmixer}, which we refer to as SC-TGN (shown in Figure \ref{fig:fig1}b) and SC-GraphMixer, respectively, with SC standing for supply chain.

\paragraph{Inventory module.}
The basic idea of our inventory module is that it explicitly represents each firm's inventory of products, and it adds and subtracts from the inventory based on products bought and consumed, respectively. 
Let $\inv{i} \in \mathbb{R}_{+}^{m}$ represent firm $i$'s inventory at time $t$.
We compute the total amount of product $p$ bought by firm $i$ at time $t$ as 
\begin{align}
    \textrm{buy}(i, p, t) &= \sum_{e(s, i, p, t) \in \mathcal{E}} \textrm{amt}(s, i, p, t), \label{eqn:buy}
\end{align}
where $\textrm{amt}(s, i, p, t)$ represents the amount of the product in the transaction indexed by $s, i, p, t$.
The amount bought per product is computed directly from the data, since we observe the products that a firm buys through its transactions.
On the other hand, we cannot observe the products that a firm \textit{consumes} from their inventory; only the finished products that they supply to others.
So, we need to learn the function mapping from products externally supplied to products internally consumed from the inventory.
We estimate the total amount of product $p$ consumed by firm $i$ at time $t$ as
\begin{align}
    \textrm{cons}(i, p, t) &= \sum_{e(i, b, p_s, t) \in \mathcal{E}} \alpha_{p_{s}p} \cdot \textrm{amt}(i, b, p_s, t), \label{eqn:consume}
\end{align}
where $\alpha_{p_{s}p} \in \mathbb{R}_+$ is a learned attention weight representing how much of product $p$ is needed to make one unit of the product supplied, $p_s$.
In other words, each product \textit{attends to} its parts.
Finally, let $\mathbf{b}_i^{(t)}$ represent the vector over all products of amount bought by firm $i$ at time $t$, and let $\mathbf{c}_i^{(t)}$ be defined analogously for consumption.
Then the firm's updated inventory is
\begin{align}
    \mathbf{x}_i^{(t+1)} &= \max(0, \inv{i} + \mathbf{b}_i^{(t)} - \mathbf{c}_i^{(t)}).
\end{align}
We take the elementwise max with 0 to ensure that the inventory stays non-negative, but we also penalize whenever consumption exceeds the current inventory amounts \eqref{eqn:inv-loss}.

\textit{Attention weights.}
If the inventory module is standalone, then we directly learn the pairwise weights $\alpha_{p_{1}p_{2}}$ for all product pairs $p_1, p_2$.
If the inventory module has access to product embeddings (e.g., from a GNN), we can use product embedding $\mathbf{z}_p \in \mathbb{R}^d$ to inform the attention weights as 
\begin{align}
    \alpha_{p_{1}p_{2}} &= \textrm{ReLU}(\mathbf{z}_{p_1} \mathbf{W}_{\textrm{att}} \mathbf{z}_{p_2} + \nu_{p_{1}p_{2}}), \label{eqn:bilinear}
\end{align}
where $\mathbf{W}_{\textrm{att}} \in \mathbb{R}^{d \times d}$ and $\nu_{p_{1}p_{2}} \in \mathbb{R}$ are learned parameters, and we apply ReLU to ensure that the attention weights are non-negative.
Compared to directly learning the attention weights, here we treat $\mathbf{z}_{p_1} \mathbf{W}_{\textrm{att}} \mathbf{z}_{p_2}$ as the base rate and $\nu_{p_{1}p_{2}}$ as adjustments, which we encourage to be small in magnitude with $L_2$ regularization \eqref{eqn:inv-loss}.
By using the embeddings to form the base rate, instead of learning each pair independently, we can share information across product pairs, which is especially useful given sparse real-world data where we rarely observe most pairs.

\textit{Inventory loss.} 
To train the inventory module, we introduce a special loss function.
For a given firm $i$ at time $t$, its inventory loss is
\begin{align}
    \ell_{\textrm{inv}}(i,t) = \lambda_{\textrm{debt}} \sum_{p \in [m]} &\max(0, \textrm{cons}(i,p,t)- \inv{i}[p])\\
    &- \lambda_{\textrm{cons}} \sum_{p \in [m]} \textrm{cons}(i,p,t). \nonumber
\end{align}
That is, we penalize inventory debt, i.e., whenever consumption exceeds the current inventory, but otherwise reward consumption.
We penalize inventory debt since firms should not be able to consume products that they never received.
Furthermore, penalizing inventory debt results in \textit{sparse} attention weights, since for most firms, we do not observe it buying most products, so for any of those products that it does not buy, it would prefer to never consume those products since it would immediately go into inventory debt if so.
On the other hand, we need the consumption reward in order to prevent trivial solutions.
Without it, the model could learn $\alpha_{p_1,p_2} \approx 0$ for all product pairs, and it would never experience inventory debt. 
We use hyperparameters $\lambda_{\textrm{debt}}$ and $\lambda_{\textrm{cons}}$ to control the relative weight between penalizing inventory debt and rewarding consumption, and in practice, we find that choosing $\lambda_{\textrm{debt}}$ around $25\%$ larger than $\lambda_{\textrm{cons}}$ works well (Table \ref{tab:hyperparameters}).
All together, the inventory loss is
\begin{align}
    \ell_{\textrm{inv}}(t) &= \frac{1}{n} \sum_{i \in [n]} \ell_{\textrm{inv}}(i, t) + \lambda_{L_2} \sqrt{\sum_{p_1,p_2 \in [m]} \nu_{p_{1}p_{2}}^2}. \label{eqn:inv-loss}
\end{align}

\paragraph{SC-TGN.}
The first GNN we explore is Temporal Graph Network (TGN) \citep{rossi2020tgn}, which is one of the most established GNNs for dynamic link prediction, outperforming other models in the Temporal Graph Benchmark \citep{huang2023tgb}.
In our work, we have extended TGN to SC-TGN, by enabling it to (1) perform message passing over \textit{hypergraphs}, since we represent each transaction as an edge between three nodes, (2) predict edge weights (i.e., transaction amounts) in addition to edge existence. 
We also modified TGN in other ways that improved performance; we document these changes in Appendix \ref{sec:tgn-details}.
In SC-TGN, each node $i$ has a time-varying memory $\mem{i}$.
Each transaction $e(s, b, p, t)$ sends three messages: one each to the supplier $s$, buyer $b$, and product $p$.
At the end of each timestep, each node aggregates the messages it received and updates its memory, using a recurrent neural network (RNN).
Then, to produce node embedding $\emb{i}$, we apply a GNN to the node memories, so that nodes can also learn from their neighbors' memories (this is useful for preventing staleness, if a node has not had a transaction in a while).

\paragraph{SC-GraphMixer.}
We also explore GraphMixer \citep{cong2023graphmixer}, a recent model that showed that temporal GNNs do not always need complicated architectures, such as RNNs or self-attention (both of which are used by TGN), and strong performance can sometimes be achieved with simpler models that only rely on multi-layer perceptrons (MLPs).
We similarly extend GraphMixer to SC-GraphMixer, so that it can handle hypergraphs and predict edge weight in addition to edge existence (Appendix \ref{sec:graphmixer-details}).
In SC-GraphMixer, each node's embedding $\emb{i}$ is a concatenation of its node encoding and link encoding.
The node encoding is simply a sum of the node's features and mean-pooling over its 1-hop neighbors' features.
The link encoding summarizes the recent edges that the node participated in, by applying an MLP to the time encodings and features of the recent edges.

\paragraph{Decoder.}
Both models, SC-TGN and SC-GraphMixer, use the same decoder architecture, and we use the same architecture (although separate decoders) for predicting edge existence and weight.
We model these as a two-step process: first, predicting whether an edge exists; second, conditioned on the edge existing, predicting its weight.
The decoder is a two-layer MLP over the concatenated supplier firm's, buyer firm's, and product's embeddings, producing $\hat{y} \in \mathbb{R}$:
\begin{align}
    \hat{y}(s,b,p,t) &= \textrm{MLP}([\emb{s} | \emb{b} | \emb{p}]). \label{eqn:decoder}
\end{align}
Real values are natural for both prediction tasks, since for edge existence, we apply a softmax over alternatives and use cross-entropy loss to evaluate the probability of the positive transaction compared to negative samples, and for edge weight, we apply log-scaling and standard scaling to the transaction amounts so negative predicted amounts are valid.

When the inventory module is attached, we may also allow it to inform edge prediction.
For edge existence, the inventory module \textit{penalizes} impossible transactions, i.e., some $(s,b,p,t)$ where supplier $s$ does not have the parts required to make product $p$ in its inventory at time $t$, and subtracts the penalty \eqref{eqn:inv-penalty} from the model's original $\hat{y}$ \eqref{eqn:decoder}.
For edge weight, the inventory module computes the maximum amount of product $p$ that supplier $s$ could produce, based on its current inventory, and \textit{caps} $\hat{y}$ based on the computed maximum \eqref{eqn:inv-cap}.
Thus, as shown in Figure \ref{fig:fig1}b, the inventory module and GNN help each other, with the GNN's embeddings informing the inventory module's attention weights, and the inventory module's penalties and caps affecting future edge prediction.

\subsection{Model training and evaluation}
\label{sec:model_training}
\paragraph{Learning production functions.}
Our model does not have access to any production functions during training, but the goal is for the model to learn production functions via the inventory module and its specialized loss. 
For each product $p$, we have the set of its attention weights $\{\alpha_{pp_{1}}, \alpha_{pp_{2}}, \cdots, \alpha_{pp_{m}}\}$, and we compute the ranking over all products from highest to lowest weight.
We compare this ranking to the product's ground-truth parts and use average precision \eqref{eqn:aveprec} to quantify performance.
Then, we compute the mean average precision (MAP) over all products for which we have ground-truth parts.

\paragraph{Edge existence and negative sampling.}
We perform negative sampling such that each positive transaction $e(s, b, p, t)$ is paired with a set of negative transactions that did not actually occur at time $t$.
Following Temporal Graph Benchmark \citep{huang2023tgb}, we sample two types of hard negatives: first, randomly perturbing one of the three nodes; second, sampling a \textit{historical} negative, meaning a transaction that appeared in training but not at time $t$.
For each positive transaction, we sample 9 perturbation negatives and 9 historical negatives.
To evaluate performance, we use mean reciprocal rank (MRR), which evaluates the rank of the positive transaction among the negatives \eqref{eqn:mrr}.
However, MRR is non-differentiable, so during training, we use softmax cross-entropy as a strong proxy loss for MRR \citep{bruch2019rank}.

\paragraph{Edge weight.}
Since we model edge prediction as a two-stage process, we only predict edge weight (i.e., transaction amount) conditioned on the edge existing.
Thus, training and test edge weight prediction is simple: we only consider the positive transactions and we compare the model's predicted amount to the true amount using root mean squared error (RMSE) \eqref{eqn:rmse}. 
Unlike MRR, RMSE is differentiable, so we also use it in the model loss during training.
\section{Supply Chains Data}

\subsection{Real-world supply chains data} \label{sec:hitachi-data}

We acquired transactions data from TradeSparq, a third-party data provider which aggregates data from authorized government sources across 60+ countries.
Their data sources include bills of lading, receipts of reported transactions, and customs declarations.
Each product is described with a Harmonized System (HS) code, an internationally recognized system for classifying products.
Along with HS codes, the transactions data also includes the supplier firm, buyer firm, timestamp of the transaction, cost in USD, and more.
Using the TradeSparq API,
we constructed two datasets:
\begin{enumerate}
    \item Our \texttt{Tesla} dataset focuses on electric vehicles (EV) and EV parts supplied by Tesla. 
    We identified Tesla EV makers, their direct suppliers and buyers, and their suppliers' suppliers, and included all transactions between these firms from January 1, 2019, to December 31, 2022.
    \item Our industrial equipment dataset (\texttt{IED}) focuses on microscopes along with other specialized analytical and inspection equipment and their manufacturers. 
    We included makers of these products and their direct suppliers and buyers, and included all transactions between these firms in 2023. 
    For this dataset, we have also estimated the ground-truth parts of microscopes in terms of their HS codes, which we use to test our inventory module's ability to infer production functions from transactions.
\end{enumerate}

\subsection{Supply chains simulator, \texttt{SupplySim}} \label{sec:supply-sim}
\begin{figure}
    \centering
    \includegraphics[width=\linewidth]{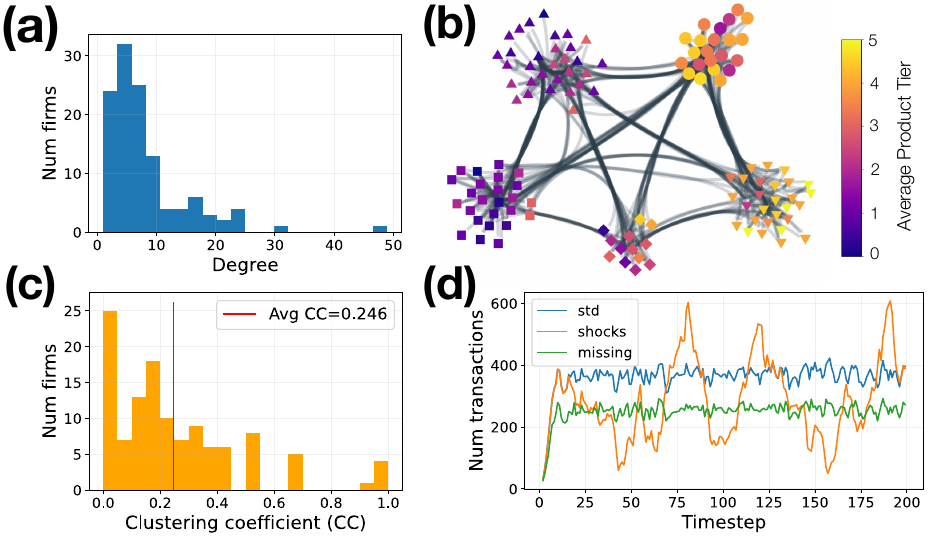}
    \caption{\texttt{SupplySim} generates data matching real data on key characteristics: \textbf{(a)} power law degree distribution, \textbf{(b)} community structure, \textbf{(c)} low clustering, \textbf{(d)} time-varying transactions, with possible shocks or missing data.}
    \label{fig:synthetic}
\end{figure}
\begin{table*}[t]
    \centering
    \small
    \begin{tabular}{p{3.7cm}|p{2cm}|p{2cm}|p{2cm}||p{2cm}}
        & \texttt{SS}-std & \texttt{SS}-shocks & \texttt{SS}-missing & \texttt{IED} \\
        \hline
        Random baseline & 0.124 (0.009) & 0.124 (0.009) & 0.124 (0.009) & 0.060 (0.002) \\
        Temporal correlations & 0.745 & 0.653 & 0.706 & 0.128 \\
        PMI & 0.602 & 0.602 & 0.606 & 0.175 \\
        node2vec & 0.280 & 0.280 & 0.287 & 0.127 \\
        Inventory module (direct) & 0.771 (0.005) & 0.770 (0.006) & 0.744 (0.006) & 0.143 (0.004) \\
        Inventory module (emb) & \textbf{0.790} (0.005) & \textbf{0.778} (0.011) & \textbf{0.755} (0.007) & \textbf{0.262} (0.005) \\
    \end{tabular}
    \caption{Results for production learning, evaluated with mean average precision (MAP $\uparrow$). For the models with randomness, we report mean and standard deviation (in parentheses) over 10 seeds.}
    \label{tab:main-pl-results}
\end{table*}
The TradeSparq data offers a rare opportunity to test models on real transactions data, but it also has shortcomings: we cannot release the data, it does not include most production functions, and, even though it provides more detailed information than most supply chains datasets, its transactions are still incomplete (e.g., missing domestic transactions).
Thus, we design a simulator, \texttt{SupplySim}, that addresses these shortcomings and enables us to test the model under \textit{controlled} settings (e.g., how much data is missing).

To ensure realism, our simulator incorporates many real-world aspects of supply chains: for example, firms specialize in certain tiers of products and time-varying transactions are generated based on the commonly used ARIO model from economics \citep{hallegatte2008ario}, which describes how firms complete orders from buyers and place orders to suppliers.
We also show that our synthetic data matches real supply chain networks on key characteristics (Figure \ref{fig:synthetic}). For example, there is community structure, but fewer triangles (thus, lower average clustering coefficient), since, unlike social networks, a firm's supplier's supplier is unlikely to also be this firm's supplier \citep{fujiwara2010structure,zhao2019adaptive}.
Our synthetic data also exhibits power law degree distributions, known to appear in real networks, and time-varying transactions with possible shocks or missing data.
Below, we briefly describe the steps of our simulator, with details in Appendix \ref{sec:synthetic-details} and \ref{sec:data-stats}. 

\paragraph{Constructing the production graph, $G_{\textrm{prod}}$.}
First, we partition the products into tiers (e.g., products 0-4 in tier 0, 5-14 in tier 1, 15-24 in tier 2, etc.) and sample a 2-dimensional position for each product from Uniform($0, 1$).
For each tier, we assign the products in the tier to parts from the previous tier,
with probability proportional to the inverse distance between their positions. 
For each part-product pair, we sample $u_{io}$, the number of units of part $p_i$ needed to make one unit of product $p_o$.
The tier structure imitates tiers in real supply chains, from raw materials in the first tier to consumer products in the final tier.
The products' positions capture the product type, e.g., its industry, and assigning parts based on positions reflects how parts should be similar to their products and naturally results in commonly co-occurring parts. 

\paragraph{Constructing supplier-buyer graph.}
For each firm,
we also sample a 2-dimensional position from Uniform($0,1$), and we restrict it to two consecutive tiers, meaning it can only produce products in those tiers.
Then, for each product, we 
select its suppliers from the set of firms that are allowed to produce that product, again with probability proportional to inverse distance. 
Now, each firm has a set of products that it is supplying, which means, based on $G_{\textrm{prod}}$, we know which input parts it needs to buy.
For each pair $(b, p)$, where firm $b$ needs to buy product $p$, we assign it to a supplier of $p$ with probability proportional to the number of buyers that the supplier already has. 
This assignment mechanism, known as preferential attachment \citep{newman2001pref}, yields power law degree distributions (Figure \ref{fig:synthetic}a) known to appear in real supply chain networks \citep{fujiwara2010structure,zhao2019adaptive}.

\paragraph{Generating transactions.}
We generate transactions based on the ARIO model, an agent-based model widely used in economics to simulate propagation over supply chains \citep{hallegatte2008ario,inoue2019japan,guan2020covid}.
At each timestep of the simulation, each firm completes as many of its incomplete orders as it can until it runs out of inventory. 
At the end of the timestep, the firm places orders to its suppliers, based on what it needs to complete its remaining orders. 
Finally, the reported transactions are the completed orders in each timestep.
The simulator also keeps track of the exogenous supply for products in the first tier, which do not require parts, and of exogenous demand for products in the final tier, which are only bought by consumers and not by other firms.
By manipulating the exogenous supply, we can model shocks in the supply chain.


\section{Experiments}
We run experiments on the two real supply chain datasets and three synthetic datasets from \texttt{SupplySim}: a standard setting with high supply (``\texttt{SS}-std''), a setting with shocks to supply (``\texttt{SS}-shocks''), and a setting with missing transactions (``\texttt{SS}-missing''), where we sampled 20\% of firms uniformly at random and dropped all of their transactions (Figure \ref{fig:synthetic}d).
In all experiments, we order the transactions chronologically and split them into train (the first 70\%), validation (the following 15\%), and test (the last 15\%).
Here we describe our results on these datasets, with additional experimental details and results in Appendix \ref{sec:experiment-details}.

\subsection{Learning production functions}
\begin{table*}[t]
    \centering
    \small
    \begin{tabular}{p{2.9cm}|p{1.7cm}|p{1.7cm}|p{1.7cm}||p{1.7cm}|p{1.7cm}}
        & \texttt{SS}-std & \texttt{SS}-shocks & \texttt{SS}-missing & \texttt{Tesla} & \texttt{IED} \\
        \hline
        Edgebank (binary) & 0.174 & 0.173 & 0.175 & 0.131 & 0.164 \\
        Edgebank (count) & 0.441 & 0.415 & 0.445 & 0.189 &  0.335\\
        Static & 0.439 (0.001) & 0.392 (0.002) & 0.442 (0.001) & 0.321 (0.001)& 0.358 (0.001)   \\  
        Graph transformer & 0.431 (0.003) & 0.396 (0.024) & 0.428 (0.003) & 0.507 (0.020) & 0.613 (0.045)  \\  
        SC-TGN & 0.522 (0.003) & 0.449 (0.004) & \textbf{0.494} (0.004) & \textbf{0.820 (0.007)} &  \textbf{0.842 (0.004)}   \\
        SC-TGN+inv & \textbf{0.540} (0.003) & \textbf{0.461} (0.009) & 0.494 (0.004) & 0.818 (0.004) & 0.841 (0.008) \\
        SC-GraphMixer & 0.453 (0.005) & 0.426 (0.004) & 0.446 (0.003) & 0.690 (0.027) & 0.791 (0.009) \\
        SC-GraphMixer+inv & 0.497 (0.004) & 0.448 (0.004) & 0.446 (0.002) &  0.681 (0.014) & 0.791 (0.008)\\
    \end{tabular}
    \caption{Results for predicting existence of future edges, evaluated with mean reciprocal rank (MRR $\uparrow$). See Table \ref{tab:edge-weight} for edge weight results.
    We report mean and standard deviation (in parentheses) over 10 seeds. }
    \label{tab:main-edge-results}
\end{table*}

We try three baselines, which each compute scores for output product $p_o$ and potential input $p_i$:
\begin{enumerate}
    \item \textbf{Temporal correlations}: we expect that inputs and outputs are temporally correlated, so this method computes the maximum correlation, with possible lags, between the buying timeseries of $p_i$ and supplying timeseries of $p_o$, averaged over all firms that buy $p_i$ and supply $p_o$.
    \item \textbf{Pointwise Mutual Information (PMI)}: we expect that input-output pairs appear with greater frequency in the firm-product graph,\footnote{The firm-product graph is a static bipartite graph of firm and product nodes, where an edge between a firm and product indicates that the firm buys or supplies the product.} so this method computes the probability that a firm buys $p_i$ \textit{and} supplies $p_o$, divided by the product of their individual probabilities \eqref{eqn:pmi}.
    \item \textbf{node2vec}: we expect that inputs are close to outputs, so this method computes the cosine similarity of $p_i$ and $p_o$'s node2vec embeddings from the firm-product graph.
\end{enumerate}
The three baselines capture the usefulness of temporal information, 1-hop neighbors in the graph, and the entire graph, respectively.
In contrast, our inventory module captures both temporal and structural information.
We try two versions of the inventory module, one that learns the attention weights directly and one that uses product embeddings, as described in \eqref{eqn:bilinear}.
We also report a random baseline, which produces uniform random rankings of parts for each product.

We summarize the production learning results in Table \ref{tab:main-pl-results}.
We find that the inventory module significantly outperforms the baselines, with especially large margins on the real-world data (\texttt{IED}) and in the synthetic data, when there are shocks in supply.
In the standard synthetic data, when supply is plentiful, temporal correlations are a strong predictor since firms place orders for inputs, receive them promptly, then supply their own outputs shortly after. 
However, once there are shocks, the firm will receive inputs at different times, due to delays, and since they cannot produce their outputs until all inputs have arrived, the correlation in time of buying inputs and supplying outputs is seriously worsened.
On the other hand, the inventory module is robust to such delays, since it does not rely on similarity in timeseries; simply that an input must go into the inventory \textit{before} the output comes out.
The inventory module is also remarkably robust to missing data: the MAP only drops by 3.5 points (4.4\%) when we drop 20\% of firms in the synthetic data.
We also see that using product embeddings, instead of learning the attention weights directly, consistently helps the inventory module.

\begin{figure}[t]
    \centering
    \includegraphics[width=\linewidth]{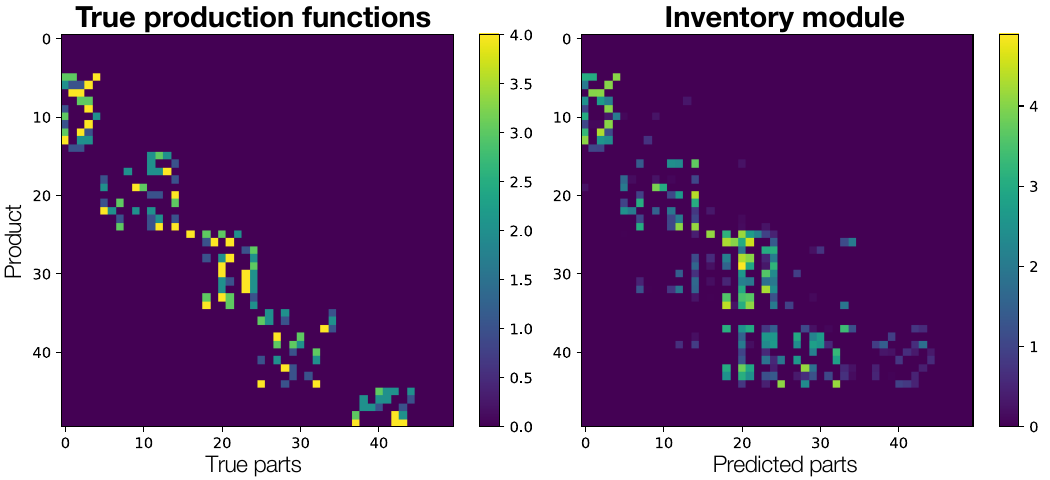}
    \caption{True production functions (left) and predictions from inventory module (right), trained on \texttt{SS}-std.}
    \label{fig:heat-map}
\end{figure}

Figure \ref{fig:heat-map} visualizes our results, showing that the inventory module effectively learns the true production functions.
The inventory module also learns attention weights of similar magnitude as the true production functions, while 
the other baselines are not comparable on magnitude, only ranking. 
While only ranking matters for MAP, magnitude is essential if using the inventory module to inform edge prediction with penalties \eqref{eqn:inv-penalty}-\eqref{eqn:inv-cap}.
Finally, in Figure \ref{fig:inv-loss-vs-map}, we show that our inventory module's loss function \eqref{eqn:inv-loss} is well-correlated with MAP, which is why it can effectively learn production functions without observing any of them.

\subsection{Predicting future edges}
For edge existence, where we seek to predict whether future transaction $(s, b, p, t)$ exists, we compare our models to the following baselines:
\begin{enumerate}
    \item \textbf{Edgebank}: ``binary'' predicts 1 if $(s, b, p, *)$ appeared before in the train set; 0 otherwise. ``count'' predicts the number of times that $(s, b, p, *)$ appeared in the train set.
    \item \textbf{Static}: learns a static vector to represent each node. 
    \item \textbf{Graph transformer}: learns a static embedding to represent each node, using the graph transformer model called UniMP from \citet{shi2021unimp}.
\end{enumerate}
While Edgebank simply memorizes the train set, it serves as a strong baseline, as the Temporal Graph Benchmark \citep{huang2023tgb} found that even the binary version outperformed some GNNs on dynamic link prediction.
The latter two baselines test two types of static node embeddings, which allow us to isolate the benefit of temporal GNNs.
For each of our models (SC-TGN and SC-GraphMixer), we try them alone and with the inventory module (+inv). 

In Table \ref{tab:main-edge-results}, we summarize results for predicting edge existence; results for predicting edge weight tell a very similar story (Table \ref{tab:edge-weight}).
First, we find that our models, SC-TGN and SC-GraphMixer, outperform the baselines on both tasks and all five datasets, by 11-62\% on edge existence and 1-13\% on edge weight.
Second, we find that SC-TGN consistently outperforms SC-GraphMixer; we hypothesize this is because of SC-TGN's sophisticated updating of node memories, while SC-GraphMixer only captures changes over time through its link encoder, which encodes the features and timestamps of the node's recent edges, and its node encoder relies on the node's static features (Appendix \ref{sec:graphmixer-details}).
Supporting this hypothesis is the fact that the static methods perform poorly, demonstrating the need for temporal GNNs for supply chains. 
Since we find that SC-TGN outperforms SC-GraphMixer, we also compared SC-TGN to ablated versions: the original TGN, to test the value of our extensions (which we document in Appendix \ref{sec:tgn-details}), and to SC-TGN where the node memory is directly used as the embedding, instead of applying a GNN to the memories.
We find that the full SC-TGN outperforms both ablations substantially, by 34-45\% and 53-100\%, respectively (Table \ref{tab:tgn-ablation}).

We also find in the synthetic data experiments, the setting with shocks is the hardest for all models.
This demonstrates how shocks complicate prediction, since shocks both delay firms' abilities to complete their orders and limit them to producing smaller amounts, thus affecting both edge existence and edge weight.
However, adding the inventory module significantly improves both SC-TGN and SC-GraphMixer's MRRs under shocks. 
In general, even though adding the inventory module introduces an additional loss so that we can learn production functions \eqref{eqn:loss}, we find that it does not hurt edge prediction performance, and in several cases, improves performance.

\subsection{Case study: generating future transactions under shocks}
In this section, we demonstrate a possible use case of our model: generating future transactions under supply chain shocks and subsequent disruptions.
For this case study, we use the synthetic data from \texttt{SS}-shocks as our ground truth, since it specifically simulates disruptions propagating over the supply chain network, due to shocks in the supply of the Tier 0 products.
We test whether our SC-TGN model (with inventory), which was trained on the first 70\% of timesteps in \texttt{SS}-shocks, can generate transactions for the first $k$ timesteps after the training data ends.
The last timestep in the training data from \texttt{SS}-shocks is $t=138$; as shown in Figure \ref{fig:synthetic}d, this timestep falls in the middle of a disruption, as the total number of transactions is falling around this timestep.

To enable this case study, we need to extend our model so that it can \textit{generate} future transactions for multiple timesteps.
We do this by utilizing its existing capabilities in predicting edge existence and edge weight:
\begin{enumerate}
    \item Predict whether a transaction exists at timestep $t$, by using the decoder \eqref{eqn:decoder} to provide a score $\hat{y}$ for each possible transaction then keeping all transactions above score threshold, i.e., $\hat{y} \geq \tau$,
    \item Conditioned on a transaction existing, predict its amount,
    \item Use the predicted transactions and amounts (as opposed to the ground-truth transactions) to update the model and inventories before timestep $t+1$.
\end{enumerate}
We use the training data to set the threshold $\tau$, by gathering the trained model's edge existence scores over all positive and negative transactions in the training data (using the negative samples described in Section~\ref{sec:model_training}) and setting $\tau$ such that the number of scores greater than or equal to $\tau$ matches the number of positive transactions in the training data. 

\begin{table}[]
    \centering
    \small
    \begin{tabular}{c|c|c}
        $t$ & AUROC (all) & AUROC (train) \\
        \hline 
        139 & 0.996 & 0.824 \\
        140 & 0.986 & 0.733 \\
        141 & 0.966 & 0.721 \\
        142 & 0.938 & 0.637 \\
        143 & 0.943 & 0.637 \\
        144 & 0.936 & 0.593 \\
        145 & 0.936 & 0.634 \\
    \end{tabular}
    \caption{Generating transactions after the end of the \texttt{SS}-shocks training data ($t=138$) and evaluating the transaction existence scores using AUROC by comparing positive transactions to all negatives (all) and train negatives (train).}
    \label{tab:case-study}
\end{table}
Using this approach, we generate transactions for $t=\{139, 140, \cdots, 145\}$, i.e., seven timesteps after the end of the training data.
At each timestep $t$, we evaluate how well the predicted edge existence scores can discriminate between the ground-truth positive transactions vs. negative transactions that did not occur at timestep $t$.
We compare to two types of negative transactions: all negatives, which keep all triplets that do not appear at time $t$, and train negatives, which keep all triplets appeared as transactions in the training data but did not appear at time $t$.
While we already evaluated the model's ability to predict edge existence in Table~\ref{tab:main-edge-results}, this is a harder setting, since we are trying to forecast transactions for several timesteps without observing any ground-truth data, which means that the model is updated with its own predicted transactions and has the risk of propagating errors over multiple timesteps.

In Table \ref{tab:case-study}, we report the area under the ROC curves (AUROC) of the predicted edge existence scores.
The AUROCs are very high when we compare the positive transactions to all negatives, and while it drops as $t$ increases, it remains above 0.93 even seven timesteps after the end of the training data.
As expected, the AUROCs are lower when we only include negatives that appeared in the training data, since those are harder to discriminate from the true positives. 
However, the AUROCs still remain decently high for the train negatives, especially for the first few timesteps.
This ability to accurately generate transactions under shocks, especially for multiple future timesteps, could help decision-makers forecast which firms or products will be affected next, enable early detection of incoming disruptions, and support sophisticated simulations of interventions.

\section{Discussion}
In this work, we have formalized temporal production graphs (TPGs), with supply chains as a canonical example, and developed a new class of GNNs that can handle TPGs.
We also release a new open-source simulator, \texttt{SupplySim}, which enables rigorous model testing and future research development in this area.
Our models successfully achieve two essential objectives---inferring production functions and predicting future edges---while preexisting GNNs focused on the latter.
Our models can be used in a wide variety of real-world scenarios, such as demand forecasting, early risk detection, and inventory optimization.
To concretely illustrate these possibilities, we conducted a case study, where we showed how the model can be used to generate future transactions under supply chain shocks and disruptions.
In future work, we hope to explore other potential use cases in supply chains, apply our model to TPGs in other domains, and develop theoretical results for our inventory module, such as establishing identifiability conditions and connecting it to causal inference.



\section*{Acknowledgements}
S.C.\ was supported in part by an NSF Graduate Research Fellowship, the Meta PhD Fellowship, and NSF award CCF-1918940. J.L.\ was supported in part by NSF awards OAC-1835598, CCF-1918940, DMS-2327709, and Stanford Data Applications Initiative. 
The authors thank Emma Pierson and members of Jure Leskovec's lab for helpful comments on the draft.
From Hitachi, we are thankful to Arnab Chakrabarti for his continuous support.

\bibliography{aaai25}


\clearpage
\appendix 
\section{Model Architecture and Evaluation} \label{sec:model-details}
In this section, we provide additional details about our SC-TGN architecture (\ref{sec:tgn-details}), our SC-GraphMixer architecture (\ref{sec:graphmixer-details}), our inventory module's link prediction penalties (\ref{sec:inv-penalties-discussion}), and our evaluation framework (\ref{sec:eval-details}).

\subsection{SC-TGN} \label{sec:tgn-details}
Here we describe SC-TGN, our extended version of TGN \citep{rossi2020tgn}, in greater detail. Like TGN, our model consists of the following modules:

\textbf{Memory.}
In SC-TGN, each node has a time-varying memory $\mem{i}$. 
The memory captures the node's state, influenced by all of the transactions we have seen for this node up to this point.
We initialize each node's memory to a \textit{learnable} vector $\mathbf{v}_i^{(0)}$. 
The node's memory is updated each time the node participates in a transaction.

\textbf{Message function.}
Each transaction, $e(s, b, p, t)$, sends a message to the three participating nodes: the supplier firm $s$, buyer firm $b$, and product $p$. 
Following the original work, we represent the raw message $\textrm{msg}_e$ as the concatenation of the nodes' memories and a time encoding:
\begin{align}
    \textrm{msg}_e &= [\mem{s} | \mem{b} | \mem{p} | \textrm{enc}(t)],
\end{align}
where $\textrm{enc}(t)$ is a temporal encoding of time $t$, implemented as a simple linear layer.
Then, the message to the supplier is $M_s(\textrm{msg}_e)$, where $M_s$ is a supplier-specific message encoder, implemented as a linear layer followed by a ReLU activation.
$M_b$ and $M_p$ are buyer-specific and product-specific message encoders, respectively, with the same architecture as $M_s$.

\textbf{Memory updater.}
First, for each firm $i$, we aggregate over the messages that it received at time $t$ to produce $\bar{msg}_i^{(t)}$. 
Following the original work, we simply take the mean over messages.
Then, we update node $i$'s memory with its aggregated message, using a recurrent neural network:
\begin{align}
    \mathbf{m}_i^{(t+1)} = RNN(\mem{i}, \bar{msg}_i^{(t)}). \label{eqn:memory-update}
\end{align}

\textbf{Embedding.}
Finally, the embedding module produces an embedding $\emb{i}$ of node $i$ at time $t$. 
Following the original work, we explore both an ``ID'' embedding, where the embedding is simply the node's memory, and embeddings that are constructed by applying a GNN to the memories.
For the GNN, we use the UniMP model \citep{shi2021unimp}, which is the model used in the TGN implementation by Temporal Graph Benchmark \citep{huang2023tgb}. UniMP is a graph transformer architecture with multi-head attention; we refer the reader to the original text for details.
The advantage of the GNN-based embedding over the ID embedding is that the GNN allows the node's embedding to adapt over time, even if the node itself has not been involved in a transaction in a while, thus avoiding the memory staleness problem \citep{kazemi2020dynamic}.
In Table \ref{tab:tgn-ablation}, we show that using a GNN to construct node embeddings from the node memories, instead of directly using the node memories, greatly improves performance of SC-TGN. 

\paragraph{New elements in SC-TGN.}
SC-TGN extends TGN in the following ways:
\begin{itemize}
    \item Hyperedges. To adapt TGN to handle hyperedges, we modify the message function so that it sends a message to three nodes, instead of two nodes. To capture the nodes' varying roles within each hyperedge (supplier, buyer, product), we also use separate message encoders $M_s$, $M_b$, and $M_p$, respectively. We also modified the decoder to predict relationships between three nodes instead of two nodes.
    \item Predicting edge weight. The original TGN only predicts edge existence, with a decoder that produces the probability of an edge existing between two nodes. Our model has two decoders: one to predict the probability of a transaction existing and, conditioned on a transaction existing, one to predict the amount of the transaction (i.e., edge weight). 
    \item Update penalty. We found it useful to regularize the memories over time, so that the memory update \eqref{eqn:memory-update} was encouraged to be small.
    \begin{align}
        \ell_{\textrm{update}} = \frac{\lambda_{\textrm{update}}}{m+n} \sum_{i \in [m+n]} ||\mathbf{m}_i^{(t+1)} - \mem{i}||_2. \label{eqn:update-loss}
    \end{align}
    \item Learnable initial memory. We initialize the node's memory to a learnable vector $\mathbf{v}_i^{(0)}$, instead of initializing all new nodes to the zero vector, which is what was done in the original TGN. Making the initial memory learnable allows the model to learn representations for nodes even before they have participated in any transactions.
    \item Training following the negative sampling scheme. In the original paper, the authors train TGN with only perturbation negatives and not historical negatives, but they test TGN on both kinds of negatives. We found that test performance increased substantially (+10 or more points in MRR) if the model could be also be trained on historical negatives.
\end{itemize}

\subsection{SC-GraphMixer} \label{sec:graphmixer-details}
Here we describe SC-GraphMixer, our extended version of GraphMixer \citep{cong2023graphmixer}, in greater detail. Like GraphMixer, our model consists of a link encoder and a node encoder:

\paragraph{Link encoder.}
The purpose of the link encoder is to capture information from recent links associated with the node.
\begin{itemize}
    \item Time encoding. Given a timestep $t$, GraphMixer encodes the timestep by projecting it to a $d$-dimensional vector.
    The time-encoding function utilizes features $\mathbf{\omega} = \{\alpha^{-(d'-1)/\beta}\}_{d'=1}^d$, where $\alpha$ and $\beta$ are predefined hyperparameters, and projects $t$ to $\cos(t \times \mathbf{\omega}) \in [-1, +1]$.
    \item To represent a given node $i$ at time $t$, GraphMixer constructs the matrix $\mathbf{T}_i(t)$.
    Each row of this matrix corresponds to a recent link of node $i$'s, where the link is represented as the concatenation of $\cos((t - t_e) \times \mathbf{\omega})$, where $t_e$ is the timestep of the link, and the link's features (in our case, the transaction amount). 
    Encoding $t - t_e$ instead of $t_e$ itself captures how recent the link was, and thus how much influence it should have over the current timestep.
    \item Then, GraphMixer applies a 1-layer MLP-mixer \citep{tolstikhin2021mlp} to $\mathbf{T}_i(t)$:
    \begin{align}
      \mathbf{H}_{\textrm{token}} &= \mathbf{T}_i(t) + \mathbf{W}^{(2)}_{\textrm{token}} GeLU(\mathbf{W}^{(2)}_{\textrm{token}} LayerNorm(\mathbf{T}_i(t))) \\
      \mathbf{H}_{\textrm{channel}} &= \mathbf{H}_{\textrm{token}} + GeLU(LayerNorm(\mathbf{H}_{\textrm{token}})W^{(1)}_{\textrm{channel}})W^{(2)}_{\textrm{channel}}.
    \end{align}
    Finally, node $i$'s link encoding at $t$, $\mathbf{t}_i(t)$, is the mean of $\mathbf{H}_{\textrm{channel}}$.
\end{itemize}

\paragraph{Node encoder.}
The purpose of the node encoder is to capture the node's identity and its features, along with its neighbors' features.
To represent a given node $i$ at time $t$, its node encoding, $\mathbf{s}_i(t)$, is
\begin{align}
    \mathbf{s}_i(t) &= \mathbf{x}_i^{\textrm{node}} + \frac{1}{|\mathcal{N}(i; t - T, t)|}\sum_{j \in \mathcal{N}(i; t - T, t)} \mathbf{x}_j^{\textrm{node}},
\end{align}
i.e., the sum of the node's own features and the mean of the node's neighbors' features, where neighbors from time $t - T$ to $t$ are included and $T$ is a predefined hyperparameter.

Finally, GraphMixer's node embedding of node $i$ at time $t$ is the concatenation of its node encoding and link encoding: $\mathbf{z}_i^{(t)} = [\mathbf{s}_i(t) | \mathbf{t}_i(t)]$.

\paragraph{New elements in SC-GraphMixer.}
SC-GraphMixer extends GraphMixer in the following ways:
\begin{itemize}
    \item Hyperedges. To adapt GraphMixer to handle hyperedges, we compute both link and node embeddings for all three nodes instead of two nodes, and the input to the decoder is the concatenation of their three embeddings \eqref{eqn:decoder}. For a product node, its recent links involve all recent transactions that it was involved in. For a firm node, its recent links involve all recent transactions where it was the supplier or the buyer. As in GraphMixer, the time encoding encodes the difference in time between when the transaction occurred and the current timestep. 
    \item Learnable node features. The original GraphMixer treats node features as per-node fixed input to models. Inspired by the memory module of TGN, we introduced \textit{learnable} node feature $\mathbf{x}_i^{\textrm{node}}$ for each node, when the dataset does not provide raw node features. The node features are randomly initialized and optimized as model parameters. 
    Introducing learnable node features also means that the static node embedding is a strictly nested version of SC-GraphMixer, making it a proper ablation of both of our models.
    \item Predicting edge weight in addition to edge existence, as described above for SC-TGN.
    \item Training following the negative sampling scheme, as described above for SC-TGN.
\end{itemize}

\subsection{Penalties from inventory module} \label{sec:inv-penalties-discussion}
As described in the main text, when our inventory module is combined with a GNN, such as SC-TGN or SC-GraphMixer, we can use the inventory module to not only learn production functions but also help with predicting future edges.
To help with predicting edge existence, the inventory module \textit{penalizes} impossible transactions, i.e., some $(s, b, p, t)$ where the supplier $s$ does not have enough parts to make product $p$.
We define the penalty as
\begin{align}
    pen(s, b, p, t) &= -\sum_{p' \in [m]} \max(0, \alpha_{pp'}-\inv{s}[p']). \label{eqn:inv-penalty}
\end{align}
Then, our model's new prediction for edge existence becomes the sum of the original $\hat{y}$ in \eqref{eqn:decoder} and the penalty \eqref{eqn:inv-penalty}.
To help with predicting edge weight, i.e., transaction amount, the inventory module computes a \textit{cap} on the maximum amount of product $p$ that supplier $s$ could produce at this time.
\begin{align}
    cap(s, b, p, t) &= \min_{p' \in [m]; \alpha_{pp'}>0} \{\frac{\inv{s}[p']}{\alpha_{pp'}}\} \label{eqn:inv-cap}
\end{align}
Then, our new predicted amount is the minimum of the original predicted amount \eqref{eqn:decoder} and the cap (after applying the same log-scaling and standard scaling to the cap that we did to all amounts).
Note that, if $\alpha_{pp'}$ matches the ground-truth production function $\mathcal{F}_p$, i.e., exactly $\alpha_{pp'}$ amount of $p'$ is needed to make one unit of $p$, and the inventory at time $t$ is correct, these penalties and caps can only help, not hurt, the prediction.
Specifically, the penalty will always be 0 for a true transaction while it can be negative for a false transaction, and the cap is guaranteed to be greater than or equal to the true transacted amount.
Since the inventory is only updated up to time $t$, but the current batch could include transactions at time $t+1$ or later, we only apply penalties or caps to transactions in time $t$, since at time $t+1$ and later, the inventory will look different.

However, even if the learned attention weights match the ground-truth production functions closely, it is not guaranteed that the inventory reflects what each firm truly has at the time of the transaction.
One minor case that we see of this is that the inventory is only updated up to time $t$, but the current batch could include transactions at time $t+1$ or later; thus, we only apply penalties and caps to transactions in time $t$, since at time $t+1$ and later, the inventory will look different.
A more serious case is when we are systematically missing transactions in the dataset, so as a result, we are underestimating the number of products that each firm has in its inventory, so penalties are too large and caps are too low.
We investigate this issue in Appendix \ref{sec:extra-link-pred} and show that, when some firms are missing from the synthetic data, the penalties and caps indeed hurt edge prediction performance, even if the attention weights are perfectly learned (we set them to the ground-truth production functions). So, it is preferable not to use the penalties and caps in those settings.
In future work, it would be useful to explore how to make the inventory module more robust to missing data, so that we do not underestimate the number of products in the firms' inventories even when transactions are missing.
For example, some smoothing could be applied such that when we observe a firm buying product $p$, we also add a bit of $p'$ to its inventory, where $p'$ is a ``similar'' product, where similarity could be defined based on co-buying patterns.

\subsection{Model training and evaluation}
\label{sec:eval-details}

In summary, our model loss consists of four components,
\begin{align}
    \mathcal{L} &= \ell_{\textrm{exist}} + \ell_{\textrm{weight}} + \ell_{\textrm{inv}} + \ell_{\textrm{update}}, \label{eqn:loss}
\end{align}
which correspond to the softmax cross-entropy loss from predicting edge existence; RMSE loss from predicting edge weight \eqref{eqn:rmse}; the inventory loss \eqref{eqn:inv-loss}; and an update loss that regularizes how much the node memories change between consecutive timesteps \eqref{eqn:update-loss}. 
Aside from $\ell_{\textrm{exist}}$, each of the other losses has hyperparameters that control their relative weight.
In Table \ref{tab:hyperparameters}, we document the hyperparameters we used in our experiments.

\paragraph{Learning production functions.}
We use mean average precision (MAP) to evaluate how well the inventory module learned the ground-truth production functions.
Let $\mathcal{P}_{p}$ represent the ground-truth set of parts for product $p$, let $\alpha_{p} = \{\alpha_{pp_1}, \alpha_{pp_2}, \cdots, \alpha_{pp_m}\}$ represent the inventory module's learned attention weights for $p$, and let $\alpha_{p}^{(k)}$ represent the product at rank $k$ in $\alpha_{p}$.
We define average precision as
\begin{align}
    \textrm{AvePr}(p) &= \frac{1}{|\mathcal{P}_{p}|} \sum_{k=1}^m \textrm{Pr@K}(\alpha_{p}, \mathcal{P}_{p}, k) \cdot \mathds{1}[\alpha_{p}^{(k)} \in \mathcal{P}_{p_o}] \label{eqn:aveprec} \\
    \textrm{Pr@K}(\alpha, \mathcal{P}, k) &= \frac{1}{k}\sum_{k'=1}^k \mathds{1}[\textrm{pos}(\alpha, k') \in \mathcal{P}]. \label{eqn:prec@k}
\end{align}
In other words, average precision takes the average of Precision@K over all ranks $k$ where there is a relevant item.
Average precision and Precision@K are both common metrics in information retrieval, but we prefer average precision here since it can account for varying numbers of relevant items over queries, which applies to our setting since the size of $\mathcal{P}_{p}$ varies across products.

\paragraph{Predicting edge existence.}
As described in the main text, we sample two types of negatives, perturbation negatives and historical negatives, to accompany each positive transaction.
Temporal Graph Benchmark \citep{huang2023tgb} also samples both types of negatives, although we modify the procedure slightly for hyperedges.
For a positive transaction $e(s, b, p, t)$, we have three types of perturbation negatives: $e(s', b, p, t)$, where a random supplier is sampled; $e(s, b', p, t)$, where a random buyer is sampled; and $e(s, b, p', t)$, where a random product is sampled.
A historical negative is a transaction $e(s', b', p', t)$ such that $s', b', p'$ appeared in the train set, but not at time $t$.
Compared to a uniform random negative, these negatives are far harder to distinguish from the positive transaction. 
For the perturbations, they match the positive transaction on two of the three nodes, requiring the model to make a distinction based on the third node.
The historical negatives are even harder: this transaction \textit{has} been observed before, but it does not occur at this timestep, requiring the model to learn time-varying properties of the graph.
To make historical negatives even harder, we sample historical negatives proportionally to their count in the train set.
For both types of negatives, we ensure that the sampled negative did not actually occur at time $t$.

We sample 9 negatives following the perturbation procedure (with 3 of each type) and 9 negatives following the historical procedure.
It is possible, though, for more than 9 negatives to be historical (if the perturbation procedure results in a historical negative) or more than 9 negatives to be perturbations (if a historical negative happens to be a one-node perturbation).
We use mean reciprocal rank (MRR) to evaluate performance for predicting edge existence.
For each positive transaction $e$ and its corresponding set of negatives $\mathcal{N}_e$, MRR computes the rank of the positive transaction's predicted score, $\hat{y}_e$, among the negative samples' predicted scores.
Following the Temporal Graph Benchmark \citep{huang2023tgb}, we compute an ``optimistic'' rank, $r_{\textrm{opt}}(e)$, where we tie-break in favor of the positive transaction, and a ``pessimistic'' rank, $r_{\textrm{pes}}(e)$, where we tie-break in favor of the negative transactions:
\begin{align}
    r_{\textrm{opt}}(e) &= \sum_{n \in \mathcal{N}_e} \mathds{1}[\hat{y}_n < \hat{y}_e] \\
    r_{\textrm{pes}}(e) &= \sum_{n \in \mathcal{N}_e} \mathds{1}[\hat{y}_n \leq \hat{y}_e].
\end{align}
Then, MRR takes the reciprocal of the average of the two ranks (+1 to avoid dividing by 0), over all positive transactions $e$ in batch $B$:
\begin{align}
    \textrm{MRR} = \frac{1}{|B|} \sum_{e \in B}\left ( \frac{r_{\textrm{opt}}(e) + r_{\textrm{pos}}(e)}{2} + 1 \right )^{-1}. \label{eqn:mrr}
\end{align}
We refer to 0.1 as the random baseline, since we sample 18 negatives in total and, if all transactions received the same score, then the MRR would be $((0 + 18)/2 + 1)^{-1} = 0.1$.
Similarly, if we assigned uniform random scores to all 19 alternatives, the expected rank of the positive is 10 (9 negatives before and 9 negatives after), so the reciprocal is 0.1.

\paragraph{Predicting edge weight.}
We preprocess transaction amounts by applying log-scaling then standard scaling, where we subtract the mean and divide by the standard deviation, so that amounts now have zero mean and variance of 1.
Scaling is useful when using amount as an edge feature, to stabilize training, and as a prediction target, to make RMSE comparable across datasets.
However, note that we use the raw amount to update the inventory module, as required in \eqref{eqn:buy}-\eqref{eqn:consume}.
Then, we compare the predicted amount $\hat{y}$ to the true amount, $y$, using root mean squared error:
\begin{align}
    RMSE = \sqrt{\frac{1}{|B|} \sum_{e \in B} (amt(e) - \hat{y}_e)^2}. \label{eqn:rmse}
\end{align}
Since RMSE is differentiable, we use it both as the training loss and evaluation metric.
\section{Supply Chains Data} \label{sec:data-details}
In this section, we provide greater detail about our real-world supply chains datasets from TradeSparq (\ref{sec:hitachi-data-details}), our simulator \texttt{SupplySim} for generating realistic supply chains data (\ref{sec:synthetic-details}), and network characteristics of our synthetic data, especially how they relate to real-world supply chain networks (\ref{sec:data-stats}).
We also provide a summary of dataset statistics in Table~\ref{tab:data-stats}.
\begin{table*}[]
    \centering
    \begin{tabular}{p{3.7cm}|p{2.8cm}|p{2.8cm}|p{2.8cm}}
        & \texttt{SS}-std & \texttt{Tesla} & \texttt{IED} \\
        \hline
        \# Product Nodes & 50 & 2,690 & 3,029 \\        
        \# Firms Nodes & 119 & 11,628 & 2,583 \\
        \# Transactions & 71646 &        581,002 & 279,712 \\
        Timespan (Days) & 198 & 1683 & 359 \\
    \end{tabular}
    \caption{Dataset statistics.}
    \label{tab:data-stats}
\end{table*}

\subsection{Real-world supply chains data} \label{sec:hitachi-data-details}
For our experiments, we constructed two large-scale, real-world supply chain datasets by compiling data from TradeSparq, a third-party data provider. We rely on a wide variety of data sources from authorized government sources across over 60 countries, including authorities like the Automated Manifest System, operated by US Customs and Border Protection, and other government customs departments. The data appear in the form of bills of lading, receipts of reported transactions, and custom declarations.
Each product being transacted is described with a Harmonized System (HS) code, where HS6 is an internationally recognized six digit number that describes products in terms of their chapters (first two digits), headings (next two digits), and subheadings (last two digits). 
For example, the HS code 840731 refers to the chapter 84 (chapter: Machinery), 07 (heading: Spark-ignition Reciprocating Piston Engines), and 31 (sub-heading: engines type not over 50CC cylinder capacity).
Beside HS code, each transaction also includes data source, the buyer firm's name and ID, the supplier firm's name and ID, a product description in free text, import and export country, cost of reported transaction, weight, and number of units, among other columns.  


We have access to the data through an API, which allows us to make queries with associated costs. 
Due to these costs, we cannot reconstruct the entire supply chain network, but we focus on certain products and grow the network outwards via a breadth-first-search-like process.
The construction begins by selecting product-specific HS codes and identifying prominent Tier-0 supplier firms involved in high-volume trading.  The process captures all transactions from Tier-0 suppliers, then aggregates their group companies to identify Tier-1 suppliers, recording all corresponding transactions. This method is repeated to determine Tier-2 suppliers by analyzing the group companies of Tier-1 firms. All transactions between Tier-1 and Tier-2 firms are also collected. This systematic approach ensures a thorough mapping of supplier interactions up to Tier-2, providing detailed insights into the supply chain network.


\paragraph{Tesla dataset.}
Following this process, we constructed real-world automotive dataset within the Tesla, Inc. group companies involved in the electric vehicle (EV) and EV parts supply chain. The analysis covered transactions from January 1, 2019, to December 31, 2022. This process started by identifying Tesla EV maker firms and capturing all their transactions with Tier-1 suppliers, i.e. Tesla's direct suppliers. Subsequently, Tier-1 suppliers were identified by aggregating Tier-1 firm groups and analyzing their transactions with Tier-2 supplier firms. This structured approach provided a comprehensive mapping of Tesla’s supply chain interactions, offering detailed insights into the dynamics and dependencies within their EV supply chain network.

\paragraph{Industrial equipment dataset.} \label{sec:sem}
This dataset focuses on microscopes along with other specialized analytical and inspection equipment and their manufacturers. Spanning the years 2022 to 2023, the dataset covers approximately 630 makers. The process began with identifying Tier-0 suppliers associated with these specialized products and capturing all relevant transactions with their buyers. Further deep-tiers were constructed by aggregating transactions at Tier-1 and Tier-2 supplier levels, providing a detailed insight into the supply chain dynamics and the extensive network of suppliers involved in the production of these critical industrial tools. 
To construct the bill of material (BoM) for microscopes, we make use of an extensive survey of direct suppliers, inquiring about their products and their own suppliers. This survey, conducted in Japanese, required translating each supplied part name from Japanese to English. We then mapped these part names to HS codes using the HTS standard and manually verified the mappings. We successfully mapped and verified 87\% of parts to HS codes, due to challenges posed by difficult and miscellaneous part names. Of the mapped parts, we could find 75\% parts within our constructed supply chain. Consequently, we were able to construct 65\% of the BoM for microscopes, which we then used to evaluate our inventory module.

\paragraph{Limitations of real-world data.}
TradeSparq data provides a rare opportunity to access transaction-level information, while most supply chain studies are done on much more aggregated data (e.g., imports and exports aggregated over country) or, at best, static firm-level networks, without time-varying transactions \citep{inoue2019japan,carvalho2019primer,guan2020covid}.
However, a key limitation of the real-world data is that we cannot release it, which complicates replicability and opportunities for future work.
Furthermore, the data itself has limitations that arise from the complexity of real-world supply chains:
 \begin{itemize}
     \item Missing transactions. While TradeSparq covers 60+ countries, many transactions are also missing, especially domestic transactions that would not be reported in customs. Some companies may also deliberately obscure or omit details of their sensitive transactions in publicly available data to protect their business interests.
     \item Missing production functions. For most firms and products, we are missing their true production functions, which we need to evaluate our inventory module.
     \item Incomplete records. Essential fields such as transaction suppliers, quantities, and amounts may be missing, making it difficult to construct a complete picture of the transaction depending on the reporting country authority.
     \item Duplicate entries can arise from multiple sources reporting the same transaction. 
     \item Transactions are reported at different times than actual production dates.
     \item Due to logistic constraints, multiple parts or products may be reported together.
\end{itemize}

\subsection{Details about \texttt{SupplySim}} \label{sec:synthetic-details}
To address the limitations of real-world supply chains data, we build a simulator, \texttt{SupplySim}, which offers the following benefits:
\begin{itemize}
    \item Unlike the real-world data, we can release \texttt{SupplySim} and our generated synthetic data, improving the reproducibility and transparency of our research, along with encouraging future research in this domain.
    \item We can test the model under complete data settings, where all transactions are observed and all necessary data per transaction is provided. This allows us to disentangle model performance from data quality.
    \item We can also test how model performance changes as data quality changes, controlling how and to what extent it changes, such as dropping 20\% of firms at random.
    \item We can inject any number of shocks, of varying size and source, and test the model's ability to predict how transactions change under the shocks. In real data, large shocks are rare and confounded by many other contemporaneous factors so it is difficult to isolate their effect (e.g., shocks during the COVID-19 pandemic).
    \item We have production functions for all products, which enables us to evaluate the inventory module, and we can, again, vary the data settings to test how robust the inventory module's performance is to missing data.
\end{itemize}
Thus, \texttt{SupplySim} addresses all the limitations of real-world supply chains data, at the cost of being synthetic, instead of real-world, data. 
To mitigate this cost, we develop \texttt{SupplySim} to be as realistic as possible: for example, we ensure that the generated networks match real-world networks on key characteristics, such as degree distribution and clustering, and we use a realistic agent-based model to generate time-varying transactions over the network.

As described in the main text, our simulator progresses in three steps: (1) constructing the production graph, $G_{\textrm{prod}}$, which describes static product-part relations, (2) assigning products to supplier firms and assigning supplier firms to buyer firms, which are also static relations, (3) generating time-varying transactions between firms.

\paragraph{Constructing the production graph, $G_{\textrm{prod}}$.}
First, we define the number of exogenous products $n_{\textrm{exog}}$, number of consumer products $n_{\textrm{consumer}}$, number of products per inner tier $n_{\textrm{tier}}$, and number of inner tiers $T$.
Then, the products are assigned to tiers as follows:
\begin{itemize}
    \item Tier 0: product 0 to product $n_{\textrm{exog}}-1$,
    \item Tier 1: product $n_{\textrm{exog}}$ to product $n_{\textrm{exog}} + n_{\textrm{tier}} - 1$,
    \item Tier 2: product $n_{\textrm{exog}} + n_{\textrm{tier}}$ to product $n_{\textrm{exog}} + 2 \cdot n_{\textrm{tier}} - 1$,
    \item $\cdots$
    \item Tier $T$: product $n_{\textrm{exog}} + (T-1) \cdot n_{\textrm{tier}}$ to product $n_{\textrm{exog}} + T \cdot n_{\textrm{tier}} - 1$,
    \item Tier $T+1$: product $n_{\textrm{exog}} + T \cdot n_{\textrm{tier}}$ to product $n_{\textrm{exog}} + T \cdot n_{\textrm{tier}} + n_{\textrm{consumer}} - 1$.
\end{itemize}
Thus, there are $n_{\textrm{exog}} + T \cdot n_{\textrm{tier}} + n_{\textrm{consumer}}$ products and $T+2$ tiers overall.
In our experiments, we use $n_{\textrm{exog}} = 5$, $n_{\textrm{consumer}} = 5$, $n_{\textrm{tier}} = 10$, and $T = 4$, resulting in 50 products.
We also sample a 2-dimensional position from Uniform(0, 1) for each product.

\begin{figure}
    \centering
    \includegraphics[width=0.8\linewidth]{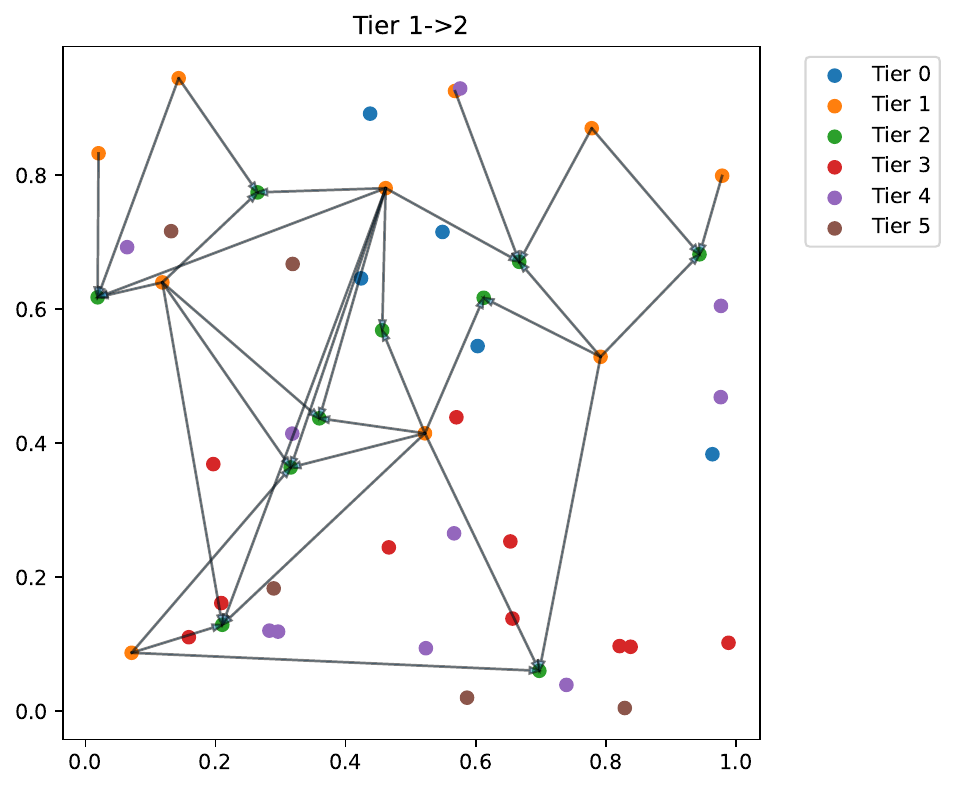}
    \caption{Visualizing products in our synthetic datasets. Each point represents the position of one of the 50 products, and points are color-coded by the product's tier. We also denote part-product relations between Tier 1 and Tier 2 products, where an arrow from product $p_1$ to product $p_2$ means that $p_1$ is required to make $p_2$. For each product, we sample its number of parts from \{1, 2, 3, 4\} uniformly, then assign its parts to the closest products in the previous tier, resulting in commonly co-occurring parts.}
    \label{fig:prod-positions}
\end{figure}
For each product, we sample its number of parts uniformly at random from a predefined range (we use 2 to 4, inclusive).
Then, we select parts for the product, where the viable parts come from the previous tier.
We assign the products to parts based on the distance between their positions, so that parts closer to the products are likelier to be chosen, with the probability proportional to $d^\gamma$, where $d$ is their distance.
In our experiments, we deterministically assign products to their closest parts for simplicity, which is essentially equivalent to choosing an extreme $\gamma$, but it would be straightforward to introduce more randomness in the selection.
In Figure \ref{fig:prod-positions}, we visualize our products' positions and the part-product relations between Tier 1 and Tier 2, demonstrating how this assignment procedure results in commonly co-occurring parts, while every product still has a unique set of parts.
Commonly co-occurring parts are realistic, such as nails and bolts, and they create possible ``similar'' parts that can be learned by GNNs.
For each pair of input part $p_i$ and output product $p_o$, we also sample $u_{io}$ uniformly at random (we use 1 to 4, inclusive), where $u_{io}$ is the number of units of the part $p_i$ needed to make one unit of the product $p_o$.

\paragraph{Constructing supplier-buyer relations.}
Now, we define the number of firms per ``group'', $n_{\textrm{group}}$, where each group is assigned a set of $n_{\textrm{consec}}$ consecutive tiers, such that the firms in the group are only allowed to supply products in this set of tiers.
We define groups and tier sets as follows:
\begin{itemize}
    \item Group 0: Tier 0 to Tier $n_{\textrm{consec}}-1$,
    \item Group 1: Tier 1 to Tier $n_{\textrm{consec}}$,
    \item Group 2: Tier 2 to Tier $n_{\textrm{consec}}+1$,
    \item $\cdots$
    \item Group $T-n_{\textrm{consec}}+2$: Tier $T-n_{\textrm{consec}}+2$ to Tier $T+1$.
\end{itemize}
In our experiments, we use $n_{\textrm{group}} = 30$ and $n_{\textrm{consec}} = 2$, resulting in 120 firms overall. 
For each firm, we also sample a 2-dimensional position from Uniform(0, 1).

Then, for each product, we sample its number of suppliers uniformly at random from a predefined range (we use 4 to 8, inclusive).
Its set of viable supplier firms are the firms within the groups that supply products in this product's tier.
For example, in our experiments, a product at Tier 2 can be supplied by firms in Groups 1 and 2 (so 60 firms). 
Similar to how we assign parts to products, we assign suppliers to products based on the closest distances between the product and firm's positions. 
Our procedure for assigning suppliers to products, where we restrict firms to consecutive tiers, results in more realistic supply chain graphs, since firms tend to specialize and we would not expect the same firm to produce products at very different tiers.
Furthermore, by restricting firms to tiers and choosing suppliers based on closest distances, we also result in firms that supply similar sets of products, which is another form of similarity that can be captured in GNN representations.

Based on the assignment of suppliers to products, and the existing production graph $G_{\textrm{prod}}$, we can now determine which input products each firm needs to buy in order to make the products it supplies.
We iterate through pairs $(b, p)$, where firm $b$ needs to buy product $p$, and select a supplier firm $s$ from all suppliers of $p$, with probability proportional to $s$'s current number of buyers.
This assignment mechanism, known as preferential attachment \citep{newman2001pref}, reflects a rich-get-richer dynamic that results in power law degree distributions, which are known to appear in real-world supply chain networks \citep{fujiwara2010structure,inoue2019japan,zhao2019adaptive}.

\paragraph{Generating time-varying transactions.}
Now, given the static graphs, we want to generate time-varying transactions between firms.
We base our simulation model on the ARIO model, an agent-based model widely used in economics to simulate propagation over supply chains \citep{hallegatte2008ario,inoue2019japan,guan2020covid}.
At any point in the simulation, we keep track of each firm's inventory $\inv{i}$; all incomplete orders $\mathcal{I}^{(t)}$, where an order is defined as $o(b, s, p, k)$, from buyer firm $b$ to supplier firm $s$ for $k$ amount of product $p$; and all orders that were newly completed at time $t$, $\mathcal{C}^{(t)}$.
Then, at each timestep $t$, we iterate through firms, with each firm $f$ taking the following actions. 
\begin{itemize}
    \item \textbf{New supply}: first, the firm adds its newly received products to its inventory, based on completed orders in $\mathcal{C}^{(t-1)}$ where it was the buyer. So, the firm is only allowed to use products from orders completed at time $t-1$ at time $t$ or later.
    \item \textbf{Production}: then, the firm goes through its incomplete orders and tries to complete as many as it can before it runs out of inventory.
    Each time the firm completes an order $o(b, f, p, k)$, it adds that order to $\mathcal{C}^{(t)}$, removes it from $\mathcal{I}^{(t)}$, and subtracts $u_{p_{i}p} \cdot k$ of input product $p_i$ from its inventory, for all inputs $p_i$ of product $p$.
    The firm goes through its incomplete orders from oldest to latest, and stops when $u_{p_{i}p} \cdot k$ exceeds what it has in its inventory.
    Thus, orders are completed first-in-first-out. This is a reasonable principle, since earlier orders should be completed earlier, but extensions could include prioritization or trying smaller, later orders after larger, earlier orders are no longer feasible. 
    \item \textbf{New demand}: finally, the firm places new orders to its suppliers. First, for each input product $p_i$, it calculates the necessary amount $k_{f,p_i}^{(t)}$ as the amount it would need to complete all of its own incomplete orders minus the amount that is already pending in its incomplete orders to its suppliers: 
    \begin{align}
        k_{f,p_i}^{(t)} &= \sum_{(b, f, p, k) \in \mathcal{I}^{(t)}} u_{p_ip} \cdot k - \sum_{(f, s, p_i, k) \in \mathcal{I}^{(t)}} k.
    \end{align}
    Then, the firm adds a new order, $o(f, s, p, k_{f,p_i}^{(t)})$ to $\mathcal{I}^{(t)}$, where $s$ is the firm's default supplier for product $p_i$ with probability $p_{\textrm{default}}$ and, with probability $1 - p_{\textrm{default}}$, we select a supplier of $p_i$ uniformly at random. We use $p_{\textrm{default}} = 0.8$, reflecting a strong level of ``stickiness'' in supplier-buyer relations, while allowing some randomness or one-off deals.
\end{itemize}

All completed orders at time $t$, $\mathcal{C}^{(t)}$, are reported as transactions for that timestep.
The simulation also requires special logic, which we describe below, for products in Tier 0, which have no parts, and products in Tier $T+1$ (the final tier), which are not used as parts for any other products and thus only bought by consumers.

\paragraph{Tier 0 products and exogenous supply.}
For each Tier 0 product $p_0$, we define time-varying exogenous supply $supply(p_0, t)$, such that any firm $f$ that supplies $p_0$ can only complete orders $o(b, f, p_0, k)$ where $k \leq supply(p_0, t)$.
In the standard synthetic data (\texttt{SS}-std), exogenous supply is plentiful, so orders for Tier 0 products are completed almost immediately.
However, we can also introduce supply shocks, which we model as follows: each Tier 0 product at each timestep $t$ has an independent probability $p_{\textrm{shock}}$ of experiencing a shock, at which point $supply(p_0, t) = supply_\textrm{shock}$.
Then, we allow the supply to recover over the subsequent timesteps, by multiplying it in each timestep by some recovery rate $r > 1$, unless it experiences another shock.

In our synthetic data with shocks, \texttt{SS-shocks}, we use $p_{\textrm{shock}} = 0.01$, resulting in 10 shocks over 200 timesteps: shocks to product 0 at times 32 and 144, product 2 at times 42 and 155, product 3 at times 51, 90, and 144, and product 4 at times 16, 58, and 131. 
We allow a recovery rate of $r = 1.25$, resulting in a recovery time of approximately 30 timesteps from the initial shock to full recovery (since we set the stable supply to 1000x of $supply_\textrm{shock}$).
In Figure \ref{fig:synthetic}d, we visualize the number of transactions over time in each synthetic data (\texttt{SS}-std, \texttt{SS}-shocks, and \texttt{SS}-missing).
While the missing firms reduce the number of transactions by a consistent fraction, the shocks dramatically disrupt the pattern of transactions, slowing them down whenever there is a shortage then trying to catch up whenever the Tier 0 products have recovered.

\paragraph{Final tier products and exogenous demand.}
For each product $p$ in the final tier (i.e., Tier $T+1$), we define time-varying exogenous demand $d(p, t)$, as well as firm-specific demand $d(f, p, t)$.
To simulate demand from consumers, with temporal patterns, we assign these products to one of three types: uniform, weekday, or weekend.
We define demand as follows:
\begin{align}
    \nu &\sim \mathcal{N}(0, 0.1) \\
    \lambda &= \begin{cases}
      1\textrm{ if type = uniform}, \\
      2\textrm{ if type = weekday and $t$ mod $7 < 5$}, \\
      0.5\textrm{ if type = weekday and $t$ mod $7 \geq 5$}, \\
      0.5\textrm{ if type = weekend and $t$ mod $7 < 5$}, \\
      2\textrm{ if type = weekend and $t$ mod $7 \geq 5$}. \\
     \end{cases} \\
    d(p, t) &= \lambda \cdot (d(p, t-1) + \nu) \\
    d(f, p, t) &= \textrm{Poisson}(d(p, t))
\end{align}
In other words, $\nu$ captures random temporal drift and $\lambda$ captures fixed weekly patterns, where uniform is the same over the week, weekday favors the first five days of the week, and weekend favors the last two days.
Then, $d(f, p, t)$ is added as an order with buyer ``consumer'' to the incomplete orders $\mathcal{I}^{(t)}$ at the beginning of each timestep $t$.
In our experiments, we allow demand to vary steadily like this, without any sharp changes, but it would be possible to simulate changes in consumer demand (e.g., due to new trends) by changing the demand schedule for final tier products.

\subsection{Characteristics of real and synthetic data}
\label{sec:data-stats}

In Figure~\ref{fig:synthetic}, we visualize several characteristics of the synthetic data generated by \texttt{SupplySim}, which match known characteristics of real-world supply chain networks.

\paragraph{Degree distribution.}
Supply chain networks are known to exhibit power law degree distributions in firm-firm networks: a few large firms dominate supply for many products \citep{fujiwara2010structure,inoue2019japan,zhao2019adaptive}.
In Figure~\ref{fig:synthetic}a, we show the degree distribution from our synthetic data, after we construct a firm-firm network, where an edge from firm $i$ to $j$ indicates that firm $i$ is a supplier of firm $j$.
As expected, the degrees in our synthetic data also follow a power law distribution due to our preferential attachment mechanism for assigning buyers to suppliers (i.e., probability proportional to how many buyers they already have).

\paragraph{Community structure.}
Supply chains are also known to have community structure, corresponding to industrial sectors or geographical regions \citep{fujiwara2010structure}.
Following \citet{fujiwara2010structure}, we quantify community structure with modularity, denoted as $Q$, which assesses the quality of a network partition into communities:
\begin{align}
    Q = \frac{1}{2E} \sum_{ij} ( A_{ij} - \frac{k_i k_j}{2E} ) \mathds{1}[c_i = c_j],
\end{align}
where $E$ is the number of edges in the network; $A_{ij}$, as the adjacency matrix, indicates whether nodes $i$ and $j$ are connected; $k_i$ is the degree of node $i$; and $\mathds{1}[c_i = c_j]$ indicates whether $i$ and $j$ are in the same community within the partition.
Intuitively, modularity compares the number of observed edges $(A_{ij})$ to the number of expected edges ($k_i k_j / 2E$) between nodes in the same community. 
Modularity can range from -0.5 to 1, and any positive value indicates community structure, meaning that there are more observed edges than expected within communities.

We try both the Louvain algorithm \citep{blondel2008fast} and a greedy algorithm \citep{clauset2004greedy}, the same used by \citet{fujiwara2010structure}, to partition our firm-firm network into communities.
Under both algorithms, we find that modularity is strongly positive, indicating community structure in our synthetic networks: $Q=0.35$ with the Louvain algorithm and $Q=0.37$ with the greedy algorithm. 
In Figure~\ref{fig:synthetic}b, we visualize the five communities found by the Louvain algorithm, and we color each firm by the average tier of its products supplied, from Tier 0 (raw materials) to Tier 5 (consumer products).
From these colors, we can see clear patterns in the communities, as the firms within them tend to specialize in products of similar tiers.
This structure arises naturally from how we constructed our synthetic data, since firms were restricted to supplying products within two consecutive tiers.

\paragraph{Clustering.}
While supply chains have community structure, they also have low clustering, since unlike social networks, a firm's supplier's supplier is unlikely to also be this firm's supplier \citep{fujiwara2010structure,zhao2019adaptive}.
In other words, we should expect fewer triangles in this network.
A node's clustering coefficient is defined as the number of observed edges between its neighbors, divided by the total possible number of edges between its neighbors, which is $k_i (k_i - 1) / 2$.
In Figure~\ref{fig:synthetic}c, we show the distribution of clustering coefficients over nodes in the firm-firm network.
Most nodes have low clustering, with an average clustering coefficient of 0.246.

\paragraph{Disassortativity.}
Another unique characteristic of supply chain networks is that they exhibit disassortativity, meaning high-degree nodes tend to be connected to low-degree nodes, instead of to other high-degree nodes \citep{fujiwara2010structure,inoue2019japan}.
Formally, assortativity is measured as the Pearson correlation between the degrees of connected nodes \citep{newman2002assortative}, where a negative correlation is interpreted as disassortativity.
We measure assortativity on our firm-firm networks and find a correlation of $r=-0.253$, indicating disassortativity.
\section{Experimental Details} \label{sec:experiment-details}
Here we provide details about our experimental set-up, including our baselines and model training / testing procedures.
Our experiments were run on a machine with 768 GB of memory and 10 GPUs (NVIDIA GeForce RTX 2080 Ti). Its operating system is Ubuntu 16.04.
Our code to run all experiments is available at \github, including necessary software libraries and versions. 

\subsection{Production learning experiments}

\paragraph{Temporal correlations.}
If product $p_i$ is a part of product $p_o$, then we expect that some firms buy $p_i$ and supply $p_o$, and it would be natural for those timeseries to be correlated, with buying slightly preceding supplying.
So, for each product pair $p_1$ and $p_2$, first we find all firms that buy $p_1$ and supply $p_2$.
If there is no such firm, then we give the pair a score of 0.
If there are such firms, then for each firm $f$, we compute the buying timeseries of $p_1$, i.e., the total amount bought by $f$ of $p_1$ on $t=1, 2, \cdots$, along with the supply timeseries of $p_2$, i.e., the total amount supplied by $f$ of $p_2$ on $t=1, 2, \cdots$.
Then, we compute the \textit{maximum} correlation between the buying and supply timeseries, allowing the supplying timeseries to lag the buying timeseries by 0 to 7 timesteps.
Finally, the score given to the pair is the average maximum correlation over firms that buy $p_1$ and supply $p_2$.

\paragraph{Pointwise Mutual Information (PMI).}
If product $p_i$ is a part of product $p_o$, then we would also expect that firms that supply $p_o$ are unusually likely to buy $p_i$, compared to the base rate of firms buying $p_i$. 
This idea is naturally quantified by PMI, where we compare the joint probability to the product of the individual probabilities.
Let $buy(p_1)$ represent whether a firm buys $p_1$ and $supply(p_2)$ represent whether a firm supplies $p_2$.
Then, we compute PMI as
\begin{align}
    PMI(p_1, p_2) &= \log \left( \frac{\Pr(buy(p_1) \wedge supply(p_2))}{\Pr(buy(p_1)) \cdot \Pr(supply(p_2))} \right), \label{eqn:pmi}
\end{align}
where the probabilities are computed as fractions of all firms.

\paragraph{node2vec.}
If product $p_i$ is a part of product $p_o$, then we would also expect that the products are close to each other in embedding space.
To test this, we train node2vec embeddings \citep{grover2016node2vec} on the firm-product graph, which is a static bipartite graph where an edge between a firm and a product indicates that the firm supplies or buys the product.
We use embedding dimension 64 for the node2vec embeddings.
Then, the score for product pair $p_1$, $p_2$ is the cosine similarity of their node2vec embeddings.

\begin{figure}
    \centering
    \includegraphics[width=\linewidth]{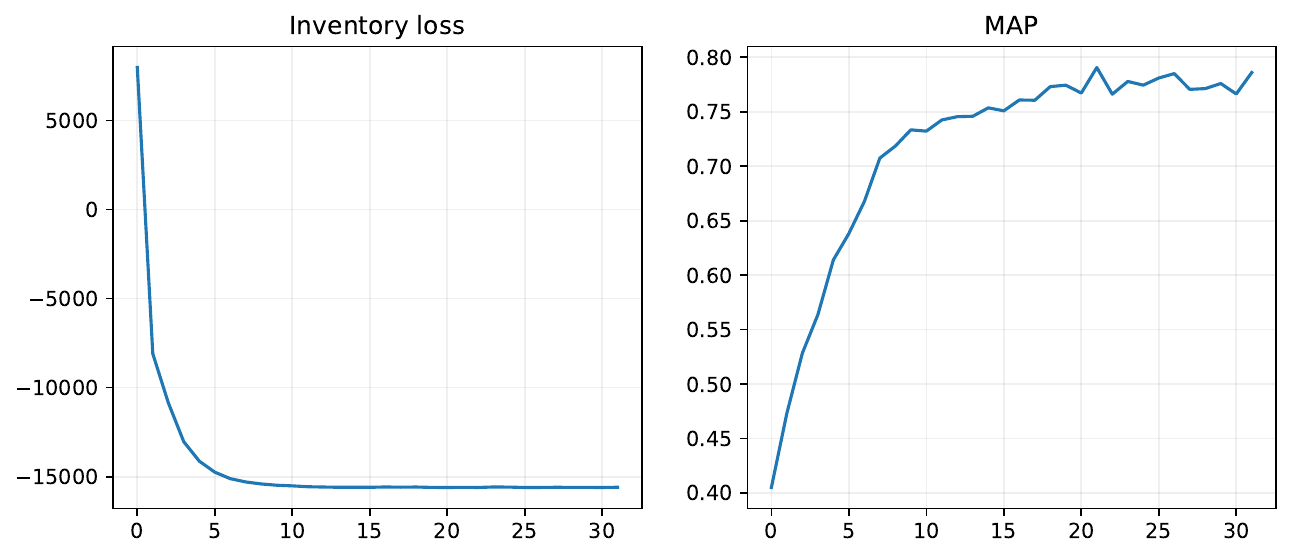}
    \caption{Comparing inventory module's loss \eqref{eqn:inv-loss} vs. MAP on ground-truth production functions, trained on \texttt{SS}-std.}
    \label{fig:inv-loss-vs-map}
\end{figure}

\paragraph{Inventory module.}
To learn production functions, we train our inventory module on the transactions data (only the train set, for consistency with the edge prediction experiments) and update it according to the inventory loss \eqref{eqn:inv-loss}.
In practice, we find it effective to choose a debt penalty $\lambda_{\textrm{debt}}$ that is 25\% larger than the consumption reward $\lambda_{\textrm{cons}}$, as reported in Table \ref{tab:hyperparameters}.
We experiment with learning the inventory module's attention weights directly and with product embeddings, following \eqref{eqn:bilinear}.
We experiment with product embeddings from node2vec, from a simple GNN trained on link prediction over the collapsed product-product graph (collapsed from the firm-product graph, so an edge from product $p_1$ to $p_2$ indicates how often a firm buys $p_1$ and supplies $p_2$), and from SC-TGN.
We find that product embeddings from node2vec help the inventory module the most on synthetic data and embeddings from the simple GNN help it the most on the microscope dataset.
In Figure \ref{fig:inv-loss-vs-map}, we also show that our inventory loss \eqref{eqn:inv-loss} is well-correlated with MAP, despite the inventory loss not having access to any ground-truth parts.

\subsection{Edge prediction experiments} \label{sec:extra-link-pred}
\begin{table*}[t]
    \centering
    \small
    \begin{tabular}{p{2.9cm}|p{1.7cm}|p{1.7cm}|p{1.7cm}||p{1.7cm}|p{1.7cm}}
        & \texttt{SS}-std & \texttt{SS}-shocks & \texttt{SS}-missing & \texttt{Tesla} & \texttt{IED} \\
        \hline
         Edgebank (avg) & 0.341 & 0.387 & 0.349 & 1.148 & 0.489 \\
        Static & 0.343 (0.008) & 0.425 (0.019) & 0.374 (0.027) & 1.011 (0.007)& 0.504 (0.018)   \\  
        Graph transformer & 0.340 (0.005) & 0.398 (0.025) & 0.361 (0.016) & 0.885 (0.024)&  0.425 (0.008)  \\  
        SC-TGN & \textbf{0.303} (0.003) & \textbf{0.359} (0.007) & 0.313 (0.002) & 0.796 (0.012)& 0.428 (0.011)  \\
        SC-TGN+inv & 0.312 (0.003) & 0.370 (0.009) & \textbf{0.312} (0.002) & 0.801 (0.015) & \textbf{0.422 (0.011)} \\
        SC-GraphMixer & 0.318 (0.003) & 0.384 (0.005) & 0.330 (0.005) & 0.774 (0.077)  & 0.457 (0.008) \\
        SC-GraphMixer+inv & 0.320 (0.004) & 0.378 (0.005) & 0.328 (0.003) & \textbf{0.767 (0.054)} & 0.454 (0.012) \\
    \end{tabular}
    \caption{Results for predicting weight of future edges (i.e., transaction amount), evaluated with root mean squared error (RMSE $\downarrow$). 
    Standard scaling was applied to amounts, so an RMSE of 1.0 should be interpreted as being off by one standard deviation.
    We report mean and standard deviation (in parentheses) over 10 seeds. }
    \label{tab:edge-weight}
\end{table*}
\paragraph{Model training.}
\begin{table*}[]
    \centering
    \begin{tabular}{c|c|c|c}
         & Synthetic data & \texttt{Tesla} & \texttt{IED} \\
        \hline
        \multicolumn{4}{c}{SC-TGN} \\
        \hline
        Memory dimension & 500 & 1000 & 1000\\
        Embedding dimension & 500 & 1000 & 1000\\
        Time dimension & 100 & 100 & 100\\
        \# neighbors for node embedding & 20 & 20 & 100\\
        Update penalty $\lambda_{\textrm{update}}$ & 1 & 1 & 1 \\
        \hline
        \multicolumn{4}{c}{SC-GraphMixer} \\
        \hline 
        \# MLPMixer layers & 2 & 2 & 2\\
        Node encoding dimension & 500 & 50 & 300 \\
        Link encoding dimension & 100 & 10 & 10 \\
        \# neighbors for node encoding & 20 & 100 & 10 \\
        \# neighbors for link encoding & 20 & 10 & 2 \\
        \hline 
        \multicolumn{4}{c}{Inventory module} \\
        \hline 
        Debt penalty $\lambda_{\textrm{debt}}$ & 5 & 5 & 5 \\
        Consumption reward $\lambda_{\textrm{cons}}$ & 4 & 4 & 4 \\
        Adjustment penalty $\lambda_{\textrm{adjust}}$ & 4 & 4 & 4 \\
        \hline
        \multicolumn{4}{c}{Training parameters} \\
        \hline
        Batch size & 30 & 30 & 100 (SC-TGN), 30 (SC-GraphMixer) \\
        Learning rate & 0.001 & 0.001 & 0.001 \\
        Max \# epochs & 100 & 100 & 100 \\
        Patience & 10 & 10 & 10 \\
    \end{tabular}
    \caption{Hyperparameters that we used in our experiments.}
    \label{tab:hyperparameters}
\end{table*}

\begin{figure*}
    \centering
    \includegraphics[width=\linewidth]{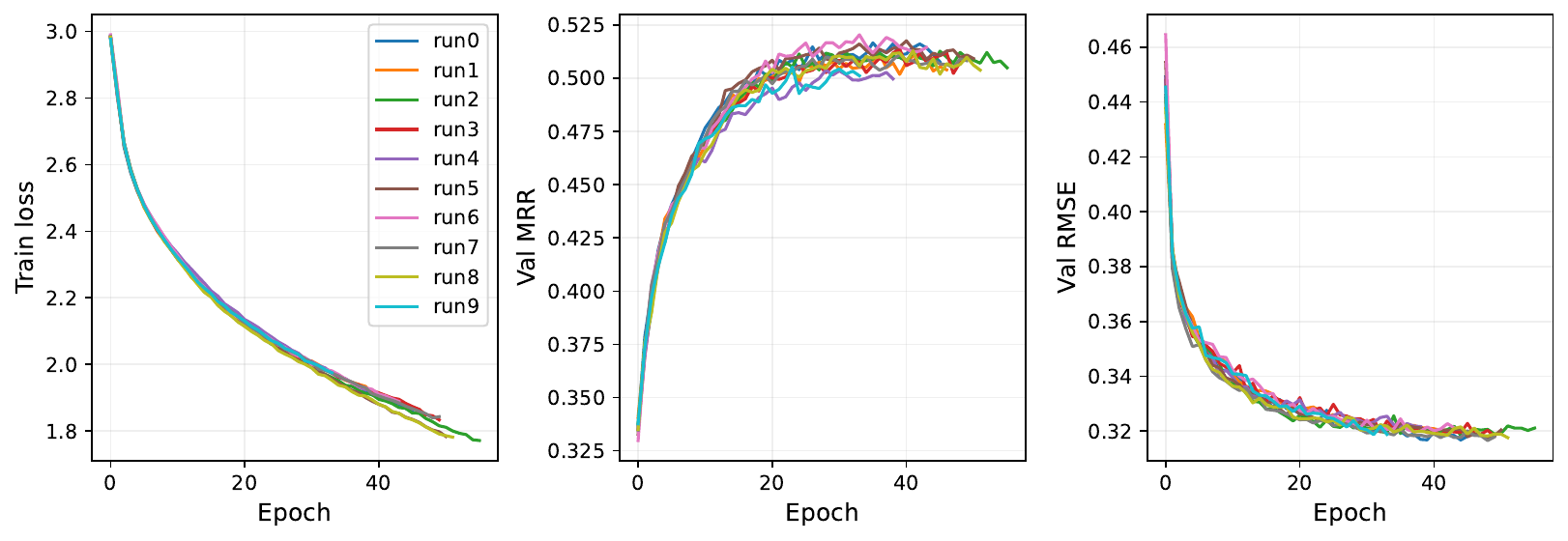}
    \caption{Performance over 10 random seeds of SC-TGN on \texttt{SS}-std.}
    \label{fig:tgn-performance-curves}
\end{figure*}
\begin{figure*}
    \centering
    \includegraphics[width=\linewidth]{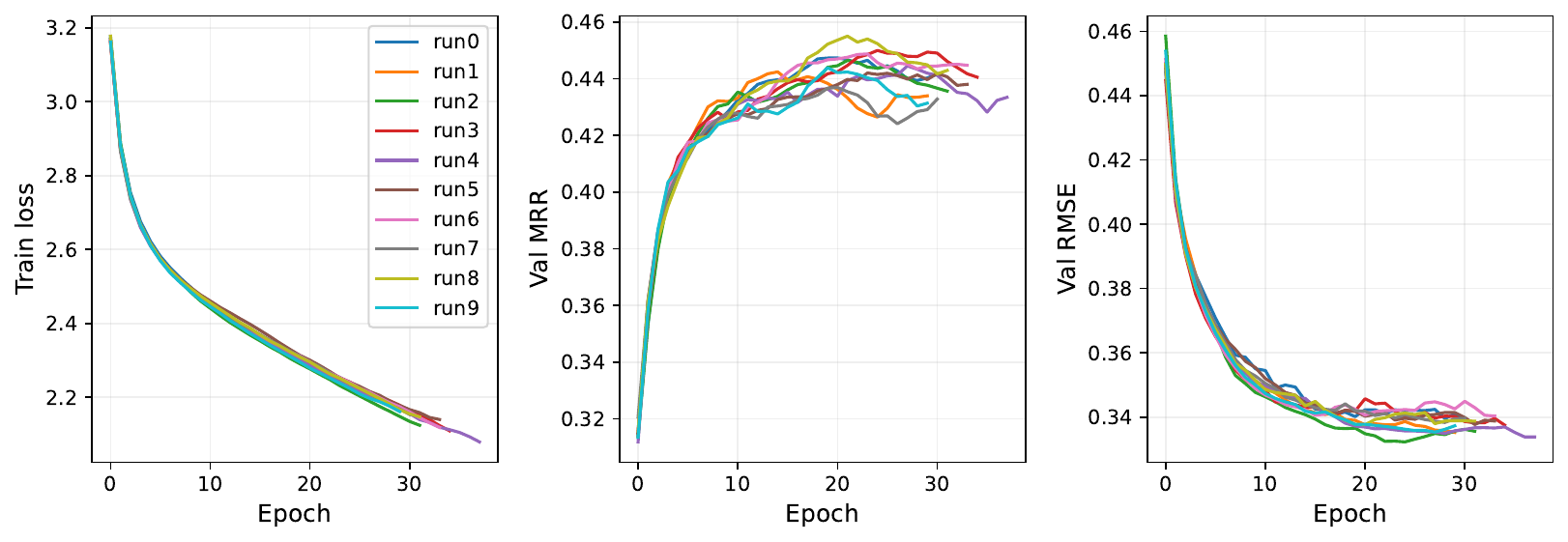}
    \caption{Performance over 10 random seeds of SC-GraphMixer on \texttt{SS}-std.}
    \label{fig:graphmixer-performance-curves}
\end{figure*}
As discussed in the main text, for all datasets, we order transactions chronologically and split them into train (70\%), validation (15\%), and test (15\%).
In these experiments, we train the models on the train set, choose hyperparameters based on the validation set, and report final results on the test set.
When training the models, we train for a maximum of 100 epochs, but we allow for early stopping, if the validation MRR does not improve for over 10 epochs.
In our main experiments (Table \ref{tab:main-edge-results}), we tested the following models: Edgebank, Static, Graph Transformer, SC-TGN, SC-TGN+inv, SC-GraphMixer, and SC-Graphmixer+inv.
In Table \ref{tab:hyperparameters}, we report the hyperparameters we used for SC-TGN and SC-GraphMixer, which we found by sweeping over the hyperparameters.
We explore different model hyperparameters, such as dimensionality (e.g., memory and embedding dimensions for SC-TGN, node feature dimensions for SC-GraphMixer) and number of neighbors for node embedding, and different training hyperparameters, such as learning rate and batch size.

For all experiments, we ran 10 trials, each with a different random seed, and reported the mean and standard deviation in test MRR and test RMSE.
As shown by the small standard deviations, performance is quite stable over random seeds, which we also show in Figures \ref{fig:tgn-performance-curves}-\ref{fig:graphmixer-performance-curves}.
These figures also show that SC-TGN and SC-Graphmixer both make steady progress in train and validation metrics over epochs: train loss decreases, validation MRR increases, and validation RMSE decreases.
It is reassuring to see that the improvements in train loss correspond well to improvements in validation MRR and RMSE, since the train loss \eqref{eqn:loss} combines multiple objectives, and, furthermore, we cannot directly train on MRR, which is non-differentiable, so instead we use softmax cross-entropy as a proxy loss.

\begin{table}[t]
    \centering
    \begin{tabular}{l|l|l}
        & \texttt{Tesla} & \texttt{IED} \\
        \hline
        TGN & 0.612 (0.009) & 0.582 (0.016)  \\
        SC-TGN (id) & 0.537 (0.021) & 0.422 (0.015) \\
        SC-TGN & \textbf{0.820} (0.007) & \textbf{0.842} (0.004) \\
    \end{tabular}
    \caption{Ablations of SC-TGN: comparing to original TGN \citep{rossi2020tgn} and SC-TGN with ID embeddings, i.e., use memory directly as embedding, instead of applying GNN to memories. We only evaluate edge existence here, with mean reciprocal rank (MRR $\uparrow$), and leave out edge weight, since the original TGN did not predict edge weight. We report mean and standard deviation (in parentheses) over 10 seeds.}
    \label{tab:tgn-ablation}
\end{table}
\paragraph{Ablations.}
We can view the Static and Graph Transformer baselines as ablations of our models, so we used the SC-TGN hyperparameters wherever applicable for fair comparison (e.g., embedding dimension, batch size).
Since SC-TGN learns the initial memory, the Static baseline is equivalent to SC-TGN if no memory updates are applied and the embedding module is ID.
The Static baseline is also a nested version of SC-GraphMixer, when we allow SC-GraphMixer to learn its initial node features.
The Graph Transformer baseline is equivalent to SC-TGN if no memory updates are applied and the embedding module is UniMP \citep{shi2021unimp}.
Since SC-TGN was our best-performing model, we also ran further ablations of SC-TGN to test the value of our extensions.
We compared it to the original TGN, where we stripped away most of the extensions discussed in Appendix \ref{sec:tgn-details}: it no longer predicted edge weight, we removed the update penalty and learnable initial memory, and we trained it only on perturbation negatives.
We also compared it to a version of SC-TGN that applied memory updates, but used ID for the embedding module, i.e., uses the node memory directly instead of applying a GNN to the memories.
As shown in Table \ref{tab:tgn-ablation}, we found that SC-TGN greatly outperformed both of these ablations as well.

\paragraph{Analysis of +inv experiments.}
\begin{table*}[t]
    \centering
    \begin{tabular}{p{3.2cm}|p{2cm}|p{2cm}|p{2cm}}
        & \texttt{SS}-std & \texttt{SS}-shocks & \texttt{SS}-missing \\
        \hline
        SC-TGN & 0.522 (0.003) & 0.449 (0.004) & \textbf{0.494} (0.004) \\
        SC-TGN+inv$^*$ & \textbf{0.548} (0.003) & \textbf{0.474} (0.003) & 0.476 (0.003) \\
        SC-GraphMixer & 0.453 (0.005) & 0.426 (0.004) & 0.446 (0.003) \\
        SC-GraphMixer+inv$^*$ & 0.477 (0.005) & 0.450 (0.005) & 0.430 (0.003) \\
    \end{tabular}
    \caption{Testing the impact of inventory module on edge existence prediction, when the inventory module is provided the ground-truth production functions. We report mean and standard deviation (in parentheses) over 10 seeds.}
    \label{tab:inv-gt}
\end{table*}
For the +inv experiments, we found that validation performance was best if we first pretrained the inventory module on only inventory loss \eqref{eqn:inv-loss}, then co-trained the inventory module and SC-TGN / SC-GraphMixer, following the joint loss \eqref{eqn:loss}.
Furthermore, in the +inv experiments, we experimented with whether to include the inventory module's edge prediction penalties or not \eqref{eqn:inv-penalty}-\eqref{eqn:inv-cap}.
Based on validation performance, we found that the penalties were useful on synthetic data, in the standard setting and under supply shocks, but not when there were missing transactions, or for the real-world datasets, so for those, we turned off the penalties.
We believe this is because, while the inventory module is robust to missing transactions when learning production functions (as shown by our production learning results in Table \ref{tab:main-pl-results}), the penalties are not robust to missing transactions, since they rely on having an approximately correct and complete inventory.
For example, if product $p_i$ is an input part to product $p_o$, and we are missing transactions where firm $f$ receives $p_i$, then the inventory module will incorrectly believe that $f$ has zero inventory of $p_i$ and penalize correct transactions where $f$ is supposed to supply $p_o$.  

To test this theory, we experimented with providing the inventory module with the \textit{ground-truth} production functions, and tested the usefulness of the edge prediction penalties on the three synthetic datasets.
We found that, on \texttt{SS}-std and \texttt{SS}-shocks, the penalties help SC-TGN and SC-GraphMixer achieve stronger performance on both edge existence and edge weight prediction, but for \texttt{SS}-missing, performance worsens (Table \ref{tab:inv-gt}).
These results show that the penalties hurting edge prediction performance are \textit{not} indicative of how well the inventory module has learned the production functions, since here we provided the ground-truth production functions.
Rather, missingness in the synthetic data causes the penalties to hurt, not help, performance. 
This analysis motivates the need to explore penalties in the future that are more robust to missingness, as discussed in Appendix \ref{sec:inv-penalties-discussion}, so that even with missing data, the inventory module can not only learn production functions, but also aid edge prediction so that the two objectives become more synergistic.

\end{document}